\documentclass[pmlr,twocolumn,10pt]{jmlr}
\mlhtrack{proceedings}
\newif\iffinal
\finaltrue
\iffinal
  \ifmlhneedspmlr
    \jmlrvolume{346}
    \jmlryear{2026}
  \fi
  \ifmlhfindings \jmlrproceedings{}{ML4H 2026 - Findings Track}\fi
  \ifmlhdemo \jmlrproceedings{}{ML4H 2026 - Demo Track}\fi
  \jmlrworkshop{Machine Learning for Health (ML4H) 2026}
\else
  \jmlrproceedings{}{Submitted to ML4H 2026: \mlhtrackname}
  \jmlrworkshop{Machine Learning for Health (ML4H) 2026}
\fi
\usepackage{booktabs}
\usepackage{multirow}
\usepackage{siunitx}
\usepackage{placeins}
\usepackage{longtable}
\usepackage{capt-of}
\hypersetup{pdftitle={From Prediction to Practice: A Task-Aware Evaluation Framework for Blood Glucose Forecasting},pdfauthor={Alireza Namazi and Heman Shakeri}}

\title[A Task-Aware Evaluation Framework for Blood Glucose Forecasting]{From Prediction to Practice: A Task-Aware Evaluation Framework for Blood Glucose Forecasting}
\author{%
\Name{Alireza Namazi} \Email{mez4em@virginia.edu}\\
\addr Department of Computer Science, University of Virginia, Charlottesville, VA, USA
\AND
\Name{Heman Shakeri} \Email{hs9hd@virginia.edu}\\
\addr School of Data Science, University of Virginia, Charlottesville, VA, USA
}
\newcommand{\glimmer}{Glimmer\textsuperscript{*}}
\begin{document}
\maketitle
\begin{abstract}
Clinical time-series forecasting is increasingly studied for decision support,
yet standard aggregate metrics can obscure failures that matter in practice.
We present a task-aware evaluation framework for blood glucose forecasting.
Forecasters warn less reliably of hypoglycemia after insulin boluses and can
predict the wrong glucose response to changes in insulin delivery.
For early warning, we evaluate four clinical cohorts. At a target of two false
alarms per patient-day, average hypoglycemia warning recall falls from 67.6\%
for other episodes to 49.5\% for episodes after an insulin bolus.
For insulin dosing support, we evaluate responses to changed insulin plans
\emph{in silico}. Several forecasters predict the wrong direction of glucose
change in more than 90\% of cases when background insulin delivery changes.
Even models that better predict responses to individual boluses often select
doses with higher simulated glucose risk. Together, these evaluations show
why average forecasting error alone is insufficient for assessing clinical
usefulness. We release the evaluation code and a simulated interventional dataset.
\end{abstract}

\paragraph*{Data and Code Availability}
Code and the simulated interventional dataset are available at
\href{https://github.com/Shakeri-Lab/benchmark-ml4h2026/tree/ml4h2026-v1.0.0}{\nolinkurl{github.com/Shakeri-Lab/benchmark-ml4h2026}}.
Clinical data (DCLP3, DCLP5, PEDAP, and IOBP2) are available under data use
agreements from the
\href{https://public.jaeb.org/datasets/diabetes}{Jaeb Center for Health Research}.
Individual clinical records are not redistributed.

\begin{samepage}
\paragraph*{Institutional Review Board (IRB)}
This study uses previously collected clinical-trial datasets distributed under
their data use agreements and simulated data; no additional IRB approval was required.
\par
\end{samepage}

\section{Introduction}
People with type~1 diabetes (T1D) depend on externally delivered insulin to
regulate blood glucose. Continuous glucose monitors (CGMs) measure interstitial
glucose at frequent intervals, enabling near-real-time feedback and
short-horizon forecasting
\citep{facchinetti2016continuous,shah2018performance,welsh2019accuracy}.
Insulin pumps deliver background insulin (basal) and discrete meal or correction
doses (boluses). Automated insulin delivery systems use CGM readings to guide
pump delivery \citep{moon2021current,lee2024shortcomings,fischer2025open}.
Too little insulin permits hyperglycemia. Too much can cause hypoglycemia.
Consensus CGM targets define low glucose as below 70~mg/dL and high glucose as
above 180~mg/dL \citep{danne2017international}. Severe hypoglycemia can cause
confusion, seizure, coma, or death. Chronic hyperglycemia contributes to
retinopathy, nephropathy, and neuropathy, while extreme hyperglycemia can cause
life-threatening metabolic crises \citep{ada2026glycemic}.

Two of the promising uses of glucose forecasting are hypoglycemia warning
and insulin dosing support. Hypoglycemia alarms may give patients time to take
carbohydrates or contact a caregiver
\citep{buckingham2010prevention,puhr2019real,harvey2012clinically}.
These alarms need sufficient lead time with few false alarms. Dosing support
requires forecasts that distinguish the consequences of different insulin plans.

We examine clinically important slices of the data: hypoglycemia episodes
soon after an insulin bolus, overnight episodes, and their complements.
Post-bolus warnings matter because bolus stacking, which occurs when a new
bolus is delivered while an earlier one is still active, can cause hypoglycemia
\citep{swan2009site,walsh2011boluscalculator}. Overnight episodes may go unnoticed
during sleep \citep{jdrf2010nocturnal}. Reporting only aggregate performance
can hide differences in the warning available in these situations.

These requirements are not captured by average prediction error alone. Most
observations occur outside acute events, and a low RMSE can coexist with poor
event prediction \citep{wolff2025blood}. Confounding in observational data can
make forecast-based dosing unsafe. When controllers deliver more insulin in
response to high or rising glucose, a forecaster may learn a spurious association
between more insulin and higher glucose. Supporting dosing decisions requires
predicting how glucose will respond when the insulin plan changes.

\begin{figure*}[t]
\centering
\includegraphics[width=\textwidth]{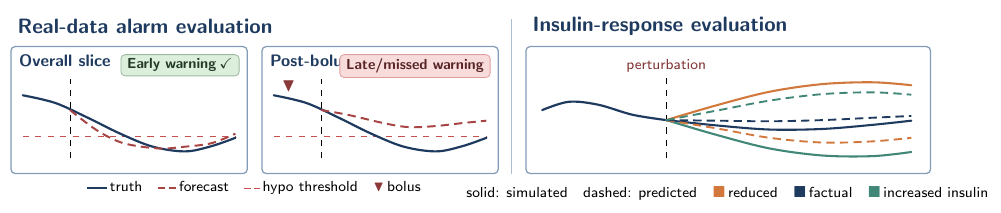}
\caption{\textbf{Two evaluations of forecast usefulness.} Schematic examples
contrast advance warning across event contexts and simulated versus predicted
responses to altered insulin. These curves illustrate the questions, not measured results.}
\label{fig:teaser}
\end{figure*}

We evaluate these two uses separately (Figure~\ref{fig:teaser}). On four clinical
cohorts, we compare recall across required warning times at common validation
false-alarm targets.
We use the FDA-accepted UVA/Padova simulator
\citep{kovatchev2009insilico,visentin2014simulator} for paired episodes that share
history and future meals but differ in future insulin. Its behavior policy is
a simulator implementation of Control-IQ, a commercial hybrid closed-loop,
model-predictive control algorithm \citep{brown2019closedloop,phillip2023consensus}.
Control-IQ adjusts basal insulin, can suspend delivery for predicted lows, and
provides automatic correction boluses alongside user-initiated meal boluses
\citep{fda2019controliq}.

The two evaluations reveal different weaknesses. Advance hypoglycemia warning
recall is generally lower in the post-bolus slice than in the non-post-bolus
slice at a stated alarm budget, with differences across warning times and cohorts.
In simulations, both deep and linear forecasters reverse basal insulin effects.
Several models predict bolus effect direction and ordering better, but still
make poor insulin dose selections.

\noindent\begin{minipage}{\columnwidth}
\textbf{Contributions.}
\begin{itemize}
\setlength{\itemsep}{1pt}
\item We combine a warning-time evaluation on clinical data with a paired
  simulation protocol for testing responses to changed insulin plans.
\item We document weaker post-bolus warnings across clinical cohorts and
  failures in insulin-response prediction and dose selection across model families.
\item We release the evaluation code and a simulated interventional dataset.
\end{itemize}
\end{minipage}\par

\section{Related Work}
Clinical safety assessment has a long history in blood glucose measurement
and forecasting. Clarke, Parkes, and Surveillance error grids assess the clinical
consequences of glucose errors
\citep{clarke1987evaluating,parkes2000new,klonoff2014surveillance}.
Forecasting studies report predictions in Parkes zones A--B as clinically
acceptable and predictions in zones C--E as clinically inaccurate
\citep{kriventsov2020diabits}. Scalar summaries and glucose-specific
losses provide complementary assessments \citep{del2012glucose,freckmann2019measures}.
The composite glucose prediction metric (CGPM) combines RMSE, temporal gain,
and glycemic-event prediction, and uses Pareto analysis to examine trade-offs
\citep{wolff2025blood}. We retain separate prediction, risk-grid, warning, and
insulin-response measurements to distinguish errors in glucose prediction,
early hypoglycemia warning, and prediction of insulin effects.

Glucose benchmarks have improved data and task standardization
\citep{sergazinov2024glucobench,metabonetbench2026}. Forecasting studies span
statistical, recurrent, Transformer, and glucose-specific approaches
\citep{karagoz2025comparative,huang2025improving}. Our focus is failure analysis,
not comprehensive model ranking. Prior work on hypoglycemia warning evaluates
warning lead times and false-alarm burden \citep{harvey2012clinically}.
It also tunes alarm thresholds separately from the glucose thresholds used to
define hypoglycemia events \citep{dassau2010suite,prendin2025alarm}.
Control-ready forecasting and digital
twin studies further motivate testing changed actions rather than only observed
trajectories \citep{controlready2024,lee2024shortcomings,herrero2026glucose}.

\section{Methods}
\subsection{Tasks, Cohorts, and Data Splits}
\label{sec:cohorts}
We study two envisioned uses of CGM forecasting: hypoglycemia warning
\citep{buckingham2010prevention,puhr2019real} and controller dosing support
\citep{moon2021current,lee2024shortcomings}. We evaluate the forecaster itself;
we do not optimize or test a complete closed-loop controller.
Each input contains past glucose and available exogenous signals on a five-minute
grid. Models forecast glucose at future five-minute steps. Forecasts for
hypoglycemia warning use recorded history. The simulator evaluation also
supplies the future basal and bolus plan.

\begin{table}[t]
\centering\small\setlength{\tabcolsep}{3pt}
\begin{tabular}{@{}lrll@{}}\toprule
Cohort & $N$ (test) & Controller & Age range\\\midrule
DCLP3 & 114 (24) & Control-IQ & 14--71 y\\
DCLP5 & 100 (20) & Control-IQ & 6--13 y\\
PEDAP & 99 (21) & Control-IQ & 2--$<$6 y\\
IOBP2 & 330 (67) & iLet & 6--79 y\\\midrule
UVA/Padova & 100 (20) & Control-IQ & Adult virtual\\\bottomrule
\end{tabular}
\caption{\textbf{Cohorts.} $N$ is the analyzed cohort size; test counts are in parentheses.}
\label{tab:cohorts}
\end{table}

The clinical trial cohorts are DCLP3 \citep{brown2019closedloop}, DCLP5
\citep{breton2020children}, PEDAP \citep{wadwa2023young}, and IOBP2 \citep{iobp2}
(Table~\ref{tab:cohorts}). PEDAP includes carbohydrate records. IOBP2 uses
frequent small automatic boluses, providing a different insulin-delivery setting.
Cohort-specific parsers align CGM,
basal, bolus, and available meal records. Appendix~\ref{app:data} reports
preprocessing, glucose distributions, and hypoglycemic-event counts.

Models are trained separately within each cohort on disjoint patient-level
70/10/20 train/validation/test splits.
We use the same split ratio for virtual subjects. Models are trained and
validated on unperturbed Control-IQ trajectories. Paired episodes with
perturbations to Control-IQ insulin plans are generated only for held-out test
subjects.

\subsection{Forecasting and Hypoglycemia Warning}
\label{sec:alarms}
\paragraph{Forecasting Metrics.}
We report glucose RMSE (mg/dL) and the percentage of forecasts in Parkes Error
Grid zones C--E (PEG C--E\%). These measure average error and risk-grid
classification, respectively. We evaluate forecasts at 30 and 60 minutes
and report the 60-minute results in the main table.

\paragraph{Observed Events and Predicted Alarms.}
An observed hypoglycemic event requires three consecutive five-minute CGM
readings below 70~mg/dL and starts at the first such reading. Three consecutive
readings at or above 70~mg/dL end the event
\citep{danne2017international}. Extraction restarts at time gaps, and low runs
already present at the start of a segment are excluded because onset is unknown.
This is a sample-based event definition, not continuous measurement of duration.

At each eligible origin, an alarm is proposed if \emph{any} predicted value in
the next 60 minutes is below a tunable alarm cutoff \citep{prendin2025alarm}.
It does not require three predicted lows. The observed-event threshold remains
70~mg/dL. After each alarm, we wait at least 60 minutes before issuing another.
Each alarm can match at most one event that begins in the following 60 minutes.
We match alarms on each patient's full timeline before calculating recall
for the post-bolus, non-post-bolus, and nocturnal slices.
A warning issued before a bolus can therefore count for a hypoglycemia episode
that starts after that bolus. Each episode receives its earliest available
unmatched warning.

\paragraph{Alarm Burden and Required Warning Time.}
For each model and cohort, we select a cutoff threshold on validation patients
to approach each target of 1, 1.5, 2, 2.5, or 3 false alarms per patient-day.
The cutoff is frozen for test, every slice, and every warning requirement.
We choose 2/day as our main reporting point. False alarms are unmatched retained
alarms, normalized by each patient's full evaluable observation time, not by
slice duration. We use 288 valid five-minute origins per observation day.

Recall at minimum warning $\ell$ is the fraction of eligible events receiving
a matched alarm at least $\ell$ minutes before onset. Every event in these
curves has a complete preceding 60-minute evaluable origin window, keeping
denominators fixed as $\ell$ varies. A warning too late for a longer requirement
does not become a false alarm. We report the full 0--60-minute curve; 0 and 5
minutes coincide on this grid because matching is strictly pre-onset.
The main table uses $\ell=15$ minutes, three sampling intervals, to distinguish
advance warning from near-onset detection. Warning-time distributions
matter because carbohydrate intake, insulin suspension, and caretaker contact
have different response times \citep{harvey2012clinically}.

\paragraph{Event Contexts and Uncertainty.}
A post-bolus hypoglycemia event is a hypoglycemia episode that starts within
30 elapsed minutes after a recorded bolus. The non-post-bolus slice contains
all remaining hypoglycemia episodes. The nocturnal slice contains episodes
starting between midnight and 06:00, a clock-time proxy for sleep
\citep{jdrf2010nocturnal}, and can overlap both bolus-defined slices.
Meal-defined slices are available only for PEDAP.
We macro-average recall over patients with events in the given slice and
false alarms/day over evaluable patients. Pointwise 95\% intervals use 2,000
patient-bootstrap draws.
Paired slice differences use patients with events in both slices.
Algorithm~\ref{alg:alarm-matching} and full results appear in the appendix.

\begin{table*}[!tp]
\centering\footnotesize
\setlength{\tabcolsep}{3pt}
\caption{\textbf{Forecasting accuracy and risk-grid metrics need not favor the same model.} 60-minute patient-level means [95\% CIs]. RMSE in mg/dL. Lower is better. Bold marks the lowest trained-model mean.}
\label{tab:std_forecasting_realdata_ci}
\begin{tabular}{@{}lcccc@{}}\toprule
& \multicolumn{2}{c}{DCLP3 ($n=24$)} & \multicolumn{2}{c}{DCLP5 ($n=20$)}\\
Forecaster & RMSE & PEG C--E\% & RMSE & PEG C--E\%\\\midrule
ARIMAX & 24.4 [22.8, 25.9] & 0.84 [0.69, 1.01] & 34.8 [32.4, 37.6] & 2.93 [2.49, 3.41] \\
DLinear & 23.1 [21.6, 24.4] & 0.70 [0.58, 0.84] & 35.0 [32.2, 38.0] & 2.42 [2.07, 2.81] \\
GRU & 22.3 [20.8, 23.6] & 0.69 [0.57, 0.85] & \textbf{30.7 [28.3, 33.3]} & 2.30 [1.86, 2.74] \\
iTransformer & 23.4 [21.9, 24.8] & 0.78 [0.66, 0.89] & 32.9 [30.1, 35.9] & 2.31 [1.91, 2.74] \\
PatchMLP & 23.5 [22.1, 24.9] & \textbf{0.60 [0.50, 0.71]} & 32.1 [29.6, 34.9] & 2.36 [1.91, 2.85] \\
AttenGluco & \textbf{22.0 [20.6, 23.2]} & 0.64 [0.55, 0.74] & 31.1 [28.6, 33.8] & \textbf{2.25 [1.87, 2.65]} \\
Glimmer\textsuperscript{*} & 22.8 [21.4, 24.2] & 0.67 [0.52, 0.89] & 32.8 [30.3, 35.7] & 2.74 [2.25, 3.25] \\
ZOH & 27.9 [26.3, 29.3] & 0.96 [0.78, 1.13] & 37.9 [35.5, 40.6] & 2.69 [2.30, 3.11] \\
\midrule
& \multicolumn{2}{c}{PEDAP ($n=21$)} & \multicolumn{2}{c}{IOBP2 ($n=67$)}\\
Forecaster & RMSE & PEG C--E\% & RMSE & PEG C--E\%\\\midrule
ARIMAX & 33.8 [31.2, 36.7] & 3.03 [2.69, 3.38] & 26.9 [25.7, 28.2] & 1.70 [1.48, 1.93] \\
DLinear & 34.3 [31.6, 37.2] & 2.66 [2.33, 3.03] & 26.4 [25.2, 27.7] & 1.33 [1.13, 1.52] \\
GRU & \textbf{29.5 [27.3, 32.0]} & 2.11 [1.84, 2.40] & \textbf{24.1 [23.0, 25.3]} & 1.31 [1.10, 1.53] \\
iTransformer & 31.4 [28.9, 34.2] & 2.23 [1.95, 2.55] & 25.4 [24.2, 26.7] & 1.23 [1.05, 1.42] \\
PatchMLP & 31.3 [28.9, 33.9] & 2.34 [2.04, 2.66] & 25.9 [24.8, 27.1] & 1.39 [1.17, 1.62] \\
AttenGluco & 30.0 [27.7, 32.6] & 2.09 [1.82, 2.38] & 24.2 [23.1, 25.4] & 1.16 [0.98, 1.35] \\
Glimmer\textsuperscript{*} & 31.1 [28.6, 33.8] & \textbf{2.04 [1.78, 2.34]} & 27.9 [26.8, 29.2] & \textbf{1.06 [0.89, 1.24]} \\
ZOH & 37.7 [34.9, 40.8] & 3.03 [2.65, 3.46] & 31.2 [29.7, 32.9] & 1.78 [1.52, 2.06] \\
\bottomrule\end{tabular}
\end{table*}

\subsection{Insulin-Response Prediction and Action Selection}
\label{sec:interventional}
We use the adult UVA/Padova cohort under a simulated Control-IQ policy. Each
episode pairs a factual plan with altered insulin plans, sharing subject,
seed, history, and future meals. We evaluate four hours after perturbation
onset, covering delayed and sustained insulin action \citep{swan2009site}.
Basal candidates multiply the future rate by
$\{0,0.5,1.5,2,2.5,3,3.5,4\}$. Bolus candidates add
$\{0,0.25,0.5,\ldots,3\}$ times the subject's mean bolus at onset.

\paragraph{Predicting Intervention Effects.}
Let $g^{\mathrm{fact}}_{t+k}$ and $g^{\mathrm{cf}}_{t+k}$ denote paired simulator
glucose, and $\hat g^{\mathrm{fact}}_{t+k}$ and $\hat g^{\mathrm{cf}}_{t+k}$
the corresponding forecasts. The true and predicted effects are
\begin{equation}
\begin{aligned}
\Delta^{\mathrm{true}}_{t+k}&=g^{\mathrm{cf}}_{t+k}-g^{\mathrm{fact}}_{t+k},\\
\Delta^{\mathrm{pred}}_{t+k}&=\hat g^{\mathrm{cf}}_{t+k}-\hat g^{\mathrm{fact}}_{t+k}.
\end{aligned}
\label{eq:effects}
\end{equation}
Effect RMSE (eRMSE) measures error in this difference. Sign agreement (SA)
measures agreement in effect direction, and Kendall's $\tau_b$
\citep{kendall1945ties} measures ordering of candidate actions. We report
30-, 60-, 120-, and 240-minute endpoints; Table~\ref{tab:cf_menu_main} uses
240 minutes, not an average across horizons. Definitions and aggregation are
in Appendix~\ref{app:simulator}.

\paragraph{Selecting Insulin Actions.}
We add the factual plan to each menu, giving nine basal and fourteen bolus
candidate labels. For action $a$, $J(a)$ is the mean Blood Glucose Risk Index
(BGRI) over the four-hour simulated glucose trajectory
\citep{kovatchev1997symmetrization}. The forecast-based cost $\hat J(a)$ is the
mean BGRI over the corresponding forecast. The oracle selects the action with
the lowest $J(a)$, and the model selects the action with the lowest $\hat J(a)$.
Action match rate (AMR) is the fraction of episodes with an exact action-label
match. Regret is the increase in simulated mean BGRI when using the model's
action $\hat a$ instead of the oracle's action $a^*$.
\begin{equation}
\begin{aligned}
a^*&=\arg\min_a J(a),\qquad \hat a=\arg\min_a\hat J(a),\\
\mathcal R&=J(\hat a)-J(a^*).
\end{aligned}
\label{eq:regret}
\end{equation}
We also score the factual Control-IQ plan and a \emph{retrospective best constant},
selected by lowest average test regret within the menu. The latter is an
optimistic test-selected reference, not a prospectively validated policy or an
upper bound on all policies. Metrics are macro-averaged over held-out virtual
subjects, with subject-bootstrapped confidence intervals. Appendix~\ref{app:simulator}
describes exact label ties and equivalent actions.

\subsection{Forecasters and Model Selection}
We compare ZOH (persistence), ARIMAX \citep{box2015time}, DLinear
\citep{zeng2023transformers}, GRU \citep{alshehri2024blood}, iTransformer
\citep{liu2023itransformer}, PatchMLP \citep{tang2025unlocking}, AttenGluco
\citep{farahmand2025attengluco}, and \glimmer. Here \glimmer{} applies the
architecture-independent Glimmer loss \citep{khamesian2025type} to PatchMLP;
it is not the original paper's full implementation. Clinical hyperparameters
are selected at 30 minutes within bounded searches around recommended settings
and reused across horizons. Simulator models are fitted from scratch and
selected only on factual simulator validation trajectories.

\paragraph{How Insulin Enters the Models.}
For neural simulator forecasts, basal and bolus channels are shifted to expose
the future plan while glucose history remains unchanged. GRU processes channels jointly.
iTransformer uses self-attention across variable tokens. AttenGluco uses
glucose queries with basal and bolus keys/values in separate cross-attention
branches, averaging the updated glucose representations. PatchMLP and
\glimmer{} use learned channel-mixing layers that modulate encoded features.
DLinear forecasts channels independently: its glucose output cannot depend on
insulin inputs, so its null intervention response is architectural.
Simulator ARIMAX predicts one glucose step from lagged glucose and insulin,
then rolls forward using predicted glucose and the known future insulin plan
(Appendix~\ref{app:models}).

\begin{table*}[!tp]
\centering\footnotesize
\setlength{\tabcolsep}{3pt}
\caption{\textbf{Aggregate recall can conceal lower post-bolus advance warning.} Minimum warning 15 minutes at a 2 FA/day validation target. Recall is percent [95\% CI]. FA/day is the achieved test rate. Bold marks the highest trained-model mean per context.}
\label{tab:hypo_gating_main}
\begin{tabular}{@{}llrccc@{}}\toprule
Cohort & Forecaster & FA/day & Overall recall & Post-bolus recall & Non-post-bolus recall\\\midrule
DCLP3 ($n=24$) & ARIMAX & 2.51 & 84.9 [80.8, 89.0] & 63.2 [40.6, 83.6] & 85.6 [81.6, 89.5] \\
 & DLinear & 2.56 & \textbf{91.6 [88.9, 94.2]} & \textbf{86.6 [66.6, 100.0]} & \textbf{91.5 [88.7, 94.1]} \\
 & GRU & 2.78 & 88.6 [84.9, 91.8] & 69.4 [47.0, 89.5] & 89.2 [85.6, 92.3] \\
 & iTransformer & 2.74 & 78.4 [72.5, 84.0] & 62.5 [36.4, 85.5] & 79.0 [73.1, 84.6] \\
 & PatchMLP & 2.54 & 87.1 [82.9, 90.9] & 73.2 [48.5, 93.9] & 87.4 [83.2, 91.1] \\
 & AttenGluco & 2.69 & 88.8 [85.9, 91.9] & 65.6 [41.8, 86.4] & 89.3 [86.6, 92.3] \\
 & Glimmer\textsuperscript{*} & 2.60 & 89.7 [86.7, 92.8] & 71.6 [47.4, 91.8] & 90.2 [87.3, 93.2] \\
 & IOB-linear & 2.66 & 90.8 [88.3, 93.3] & 82.0 [62.1, 97.4] & 91.0 [88.6, 93.6] \\
 & ZOH & 2.48 & 74.6 [68.8, 80.6] & 60.2 [35.2, 82.3] & 75.1 [69.3, 81.2] \\
\midrule
DCLP5 ($n=20$) & ARIMAX & 2.25 & 47.4 [42.6, 52.2] & 30.0 [20.9, 40.9] & 50.3 [45.7, 54.9] \\
 & DLinear & 2.27 & 52.5 [47.9, 56.9] & 31.5 [22.1, 42.0] & 56.2 [52.1, 60.0] \\
 & GRU & 2.19 & 55.1 [48.8, 61.5] & 28.8 [19.6, 39.7] & 59.7 [53.5, 65.7] \\
 & iTransformer & 2.34 & 52.4 [47.3, 57.6] & 30.5 [20.9, 41.8] & 56.1 [51.2, 61.0] \\
 & PatchMLP & 2.29 & 53.8 [48.6, 59.2] & 26.5 [17.3, 37.7] & 58.5 [53.5, 63.5] \\
 & AttenGluco & 2.28 & \textbf{60.4 [54.7, 65.9]} & \textbf{35.4 [25.0, 47.3]} & \textbf{64.9 [59.3, 69.9]} \\
 & Glimmer\textsuperscript{*} & 2.25 & 55.9 [50.1, 61.5] & 30.3 [20.2, 41.8] & 60.1 [54.7, 65.1] \\
 & IOB-linear & 2.27 & 45.3 [40.6, 50.1] & 27.5 [18.2, 38.7] & 48.3 [43.7, 52.8] \\
 & ZOH & 2.20 & 42.5 [37.7, 47.3] & 27.3 [18.0, 38.6] & 45.2 [40.6, 49.6] \\
\midrule
PEDAP ($n=21$) & ARIMAX & 1.94 & 33.0 [29.0, 37.2] & 19.7 [12.4, 29.3] & 36.2 [31.4, 41.8] \\
 & DLinear & 2.01 & 39.6 [35.0, 44.3] & 22.5 [14.1, 32.8] & 43.1 [37.8, 48.6] \\
 & GRU & 2.04 & \textbf{49.5 [44.8, 54.4]} & \textbf{27.3 [19.3, 37.0]} & \textbf{54.1 [49.3, 59.5]} \\
 & iTransformer & 2.00 & 41.5 [37.5, 45.8] & 22.5 [14.7, 32.8] & 45.5 [41.1, 49.7] \\
 & PatchMLP & 1.97 & 38.9 [35.9, 42.0] & 22.8 [14.9, 32.7] & 42.4 [39.1, 45.7] \\
 & AttenGluco & 1.95 & 47.8 [43.7, 52.3] & 27.0 [18.9, 36.2] & 52.3 [48.0, 56.8] \\
 & Glimmer\textsuperscript{*} & 1.97 & 48.2 [44.9, 51.5] & 22.9 [14.0, 33.7] & 53.1 [49.6, 56.7] \\
 & IOB-linear & 1.94 & 31.3 [27.3, 35.6] & 19.9 [12.5, 29.4] & 34.2 [29.5, 39.8] \\
 & ZOH & 1.94 & 30.0 [25.6, 34.7] & 17.9 [9.9, 28.0] & 32.9 [27.9, 38.9] \\
\midrule
IOBP2 ($n=67$) & ARIMAX & 2.21 & 79.0 [76.6, 81.3] & 72.4 [62.1, 82.3] & 79.4 [77.1, 81.7] \\
 & DLinear & 2.22 & \textbf{85.5 [83.6, 87.4]} & \textbf{81.1 [72.6, 88.9]} & \textbf{85.9 [84.0, 87.8]} \\
 & GRU & 2.22 & 79.3 [76.8, 81.9] & 71.1 [60.8, 81.3] & 79.9 [77.4, 82.3] \\
 & iTransformer & 2.19 & 77.9 [75.2, 80.4] & 65.9 [54.2, 76.7] & 78.6 [75.9, 81.1] \\
 & PatchMLP & 2.22 & 77.9 [74.8, 80.7] & 68.1 [56.3, 79.6] & 78.6 [75.8, 81.3] \\
 & AttenGluco & 2.22 & 79.9 [77.9, 82.0] & 74.2 [64.0, 84.0] & 80.5 [78.4, 82.7] \\
 & Glimmer\textsuperscript{*} & 2.26 & 79.0 [76.5, 81.6] & 71.7 [60.3, 82.4] & 79.4 [76.8, 81.8] \\
 & IOB-linear & 2.24 & 84.1 [82.0, 86.2] & 79.7 [70.8, 87.8] & 84.5 [82.4, 86.5] \\
 & ZOH & 2.13 & 66.5 [62.8, 69.9] & 59.9 [49.0, 70.2] & 66.9 [63.4, 70.3] \\
\bottomrule\end{tabular}
\end{table*}

\begin{figure*}[t]
\centering
\includegraphics[width=\textwidth]{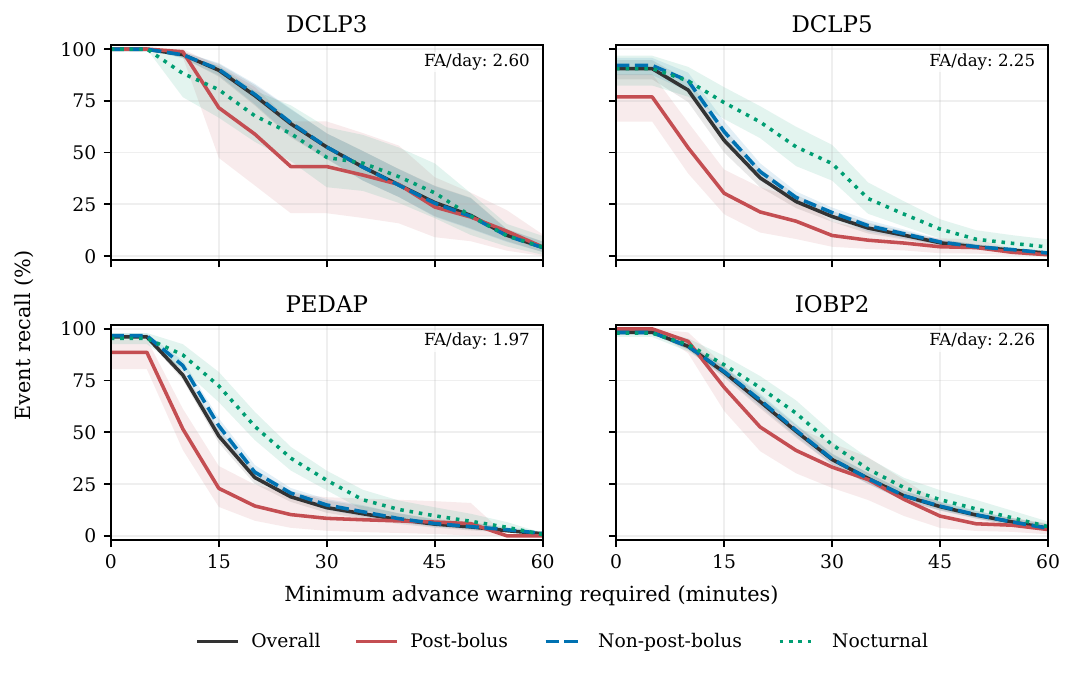}
\caption{\textbf{Required warning time exposes context differences.}
\glimmer{} event recall at a 2 FA/patient-day validation target; cutoffs are
frozen for test and all slices. Bands are pointwise 95\% patient-bootstrap
intervals. Panel annotations give achieved test FA/day.}
\label{fig:warning}
\end{figure*}
\begin{table*}[!tp]
\centering\footnotesize
\setlength{\tabcolsep}{3pt}
\caption{\textbf{Basal effects reverse; bolus responses improve.} 240-minute subject means [95\% CI]. eRMSE (mg/dL); SA: sign agreement (\%); $\tau_b$: Kendall correlation. Dashes: null response; bold: best trained-model mean.}
\label{tab:cf_menu_main}
\begin{tabular}{@{}llccc@{}}\toprule
Intervention & Forecaster & eRMSE $\downarrow$ & SA (\%) $\uparrow$ & $\tau_b$ $\uparrow$\\\midrule
Basal & ARIMAX & 141.32 [129.46, 153.69] & 4.7 [2.0, 8.3] & -0.84 [-0.88, -0.80] \\
 & DLinear & \textbf{52.10 [46.39, 58.94]} & --- & --- \\
 & GRU & 86.70 [80.25, 92.88] & \textbf{9.5 [4.8, 14.2]} & \textbf{-0.72 [-0.81, -0.65]} \\
 & iTransformer & 98.09 [90.13, 106.18] & 6.1 [2.3, 10.9] & -0.77 [-0.83, -0.71] \\
 & PatchMLP & 100.33 [93.09, 107.64] & 5.9 [2.5, 10.3] & -0.77 [-0.82, -0.72] \\
 & AttenGluco & 97.84 [91.46, 104.76] & 5.5 [2.3, 9.5] & -0.83 [-0.87, -0.78] \\
 & Glimmer\textsuperscript{*} & 99.82 [93.18, 106.72] & 5.6 [2.5, 9.7] & -0.79 [-0.84, -0.74] \\
\midrule
Bolus & ARIMAX & \textbf{25.73 [22.18, 29.03]} & \textbf{85.7 [82.1, 88.8]} & \textbf{0.67 [0.61, 0.73]} \\
 & DLinear & 37.01 [33.97, 40.56] & --- & --- \\
 & GRU & 29.92 [27.03, 32.58] & 76.2 [72.0, 80.1] & 0.49 [0.41, 0.57] \\
 & iTransformer & 38.07 [34.42, 41.46] & 47.8 [41.3, 53.9] & -0.08 [-0.19, 0.04] \\
 & PatchMLP & 31.07 [27.86, 34.35] & 76.4 [71.7, 81.0] & 0.44 [0.36, 0.53] \\
 & AttenGluco & 38.77 [35.69, 41.64] & 61.7 [55.8, 66.7] & 0.12 [0.01, 0.23] \\
 & Glimmer\textsuperscript{*} & 31.44 [28.55, 34.52] & 73.4 [68.9, 77.9] & 0.43 [0.34, 0.52] \\
\bottomrule\end{tabular}
\end{table*}

\paragraph{Simple IOB-Informed Forecaster.}
We additionally fit a three-input glucose forecaster using latest CGM, its
backward five-minute derivative, and total estimated insulin on board (IOB).
IOB uses the two-compartment remaining-insulin kernel of
\citet{lee2009mpciob}. A ridge model predicts glucose change at each lead;
training patients determine coefficients and normalization, and validation
glucose RMSE selects regularization. Its coefficients are unconstrained in sign.

\section{Results}
\label{sec:results}
\subsection{Forecasting Accuracy}
Table~\ref{tab:std_forecasting_realdata_ci} reports 60-minute RMSE and PEG C--E\%.
GRU has the lowest RMSE in three cohorts;
AttenGluco is lowest in DCLP3. The lowest risk-grid fraction varies by cohort.
These measurements do not identify a model that is uniformly best for either
warning or action selection. Appendix~\ref{app:forecasting} reports 30- and
60-minute forecasting results with 95\% intervals.

\subsection{Advance Warning at a Stated Alarm Budget}
Table~\ref{tab:hypo_gating_main} requires at least 15 minutes of warning at a
2 FA/day validation target. Actual test rates among trained forecasters range
from 1.94 to 2.78/day (median 2.25). They are not exactly matched on test.
Figure~\ref{fig:warning} shows \glimmer{} across all four cohorts. The
appendix curves cover all models.

For \glimmer, recall in the post-bolus and non-post-bolus slices is 30.3\%
and 60.1\% in DCLP5, and 22.9\% and 53.1\% in PEDAP, respectively.
Paired differences are $-29.8$
percentage points (95\% CI $-39.3$ to $-20.4$) and $-30.2$ points ($-39.2$
to $-20.6$). Its DCLP3 and IOBP2 paired estimates are also negative, but their
intervals include zero. The simple IOB forecaster shows the same direction,
including 27.5\% versus 48.3\% in DCLP5 and 19.9\% versus 34.2\% in PEDAP.
It can perform competitively with the larger models; the warning differences
are not specific to deep learning.

Across eight trained forecasters and four cohorts, all 32 paired estimates
are negative at 15 minutes, and 19 pointwise paired 95\% confidence intervals
exclude zero. Across these model--cohort comparisons, mean recall is 49.5\%
in the post-bolus slice and 67.6\% in the non-post-bolus slice.
The separate slice-level intervals do not overlap in 17 comparisons. All 32
estimates are also negative at 20 and 30 minutes, but only 24 are negative at
10 minutes. At 15 minutes across validation targets 1, 1.5, 2, 2.5, and 3/day,
the corresponding counts are 25, 30, 32, 32, and 31.

The post-bolus event counts are 32 (DCLP3), 305 (DCLP5), 424 (PEDAP), and 113
(IOBP2); DCLP3 has only 11 event-bearing patients. Sparse support limits its
precision. Nocturnal and daytime results, other alarm budgets, lead requirements,
and the complete event support appear in Appendix~\ref{app:warning-results}.

\subsection{Responses to Changed Insulin Plans}
\paragraph{Effect Direction, Magnitude, and Ordering.}
Table~\ref{tab:cf_menu_main} evaluates the 240-minute endpoint. Five deep
forecasters and ARIMAX reverse basal effects: SA is 4.7--9.5\% and Kendall's
$\tau_b$ is $-0.72$ to $-0.84$. ARIMAX has 4.7\% sign agreement,
$\tau_b=-0.84$, and eRMSE 141.32~mg/dL. Its autoregression and lower parameter
count do not prevent this failure. DLinear's basal eRMSE is lower, but its null
response makes effect direction and ranking undefined, not correct.

Bolus responses are better for several models. ARIMAX has eRMSE 25.73~mg/dL,
85.7\% sign agreement, and $\tau_b=0.67$. GRU and PatchMLP achieve about
76\% sign agreement and positive ordering. AttenGluco has weaker ordering,
and iTransformer's correlation is slightly negative. DLinear again has a null
response. Thus, basal reversal is not generic insensitivity to all insulin
inputs; discrete bolus and sustained basal perturbations expose different behavior.

\begin{table}[!tp]
\centering\small
\setlength{\tabcolsep}{3pt}
\caption{\textbf{Better bolus effect ordering does not ensure better dosing.} AMR: action match rate (\%); regret: excess BGRI cost. Bold: best trained-model mean.}
\label{tab:policy_regret_main}
\begin{tabular}{@{}llrr@{}}\toprule
Selection & Policy & AMR (\%) $\uparrow$ & Regret $\downarrow$\\\midrule
Basal & \emph{Retrospective constant} & 47.50 & 0.785 \\
 & \emph{Control-IQ} & 7.50 & 3.035 \\
\cmidrule(l){2-4}
 & ARIMAX & 3.75 & 4.936 \\
 & DLinear & 7.50 & 5.031 \\
 & GRU & 6.25 & 4.680 \\
 & iTransformer & 7.50 & 4.587 \\
 & PatchMLP & \textbf{11.25} & 4.721 \\
 & AttenGluco & 7.50 & \textbf{4.476} \\
 & Glimmer\textsuperscript{*} & 5.00 & 4.745 \\
\midrule
Bolus & \emph{Retrospective constant} & 47.50 & 1.150 \\
 & \emph{Control-IQ} & 3.75 & 3.904 \\
\cmidrule(l){2-4}
 & ARIMAX & 6.25 & 3.848 \\
 & DLinear & 6.25 & 3.904 \\
 & GRU & 8.75 & 3.062 \\
 & iTransformer & 11.25 & 3.659 \\
 & PatchMLP & 30.00 & 1.882 \\
 & AttenGluco & 13.75 & 1.960 \\
 & Glimmer\textsuperscript{*} & \textbf{31.25} & \textbf{1.715} \\
\bottomrule\end{tabular}
\end{table}

\paragraph{Selecting an Action.}
Table~\ref{tab:policy_regret_main} shows that better bolus effect prediction
does not ensure good selection: ARIMAX matches the oracle in 6.25\% of episodes,
with regret 3.848. \glimmer{} has the best trained-model bolus match rate
(31.25\%) and regret (1.715), but still trails the retrospective best constant
(47.5\%, regret 1.150). Basal selection is poor across trained models;
the highest trained-model match rate is PatchMLP's 11.25\%.

The constant is selected retrospectively within the test menu, so its advantage
is optimistic and specific to this objective. The four-hour horizon and absence
of a terminal penalty for insulin still active afterward may favor large doses
by omitting later hypoglycemia risk. This explanation requires further validation.
These results are not dosing recommendations or evidence that Control-IQ
clinically underdoses. The simulator oracle knows future meals and can evaluate
every candidate dose using information unavailable to a real-time controller.

To test whether the largest insulin perturbations accounted for the high
regret, we restricted both the models and the oracle to scales of at most
$2\times$. Poor dose selection persisted. ARIMAX basal/bolus regrets are
3.442/3.039, compared with
0.219/0.836 for the restricted-menu retrospective constant. Deep-model regrets
range from 2.962--3.237 (basal) and 1.431--2.997 (bolus).
For bolus actions, the cap applies to the added insulin relative to the
reference dose.
Appendix~\ref{app:sim-results} provides uncertainty, action-confusion matrices,
and additional endpoint results.

\section{Discussion}
\paragraph{Why Insulin Effects May Be Reversed.}
In the observational simulator episodes, higher insulin delivery is
correlated with higher subsequent glucose ($r=0.451$ basal, $0.458$ bolus).
The controller responds to high or rising glucose with more insulin. Under controlled
perturbations, insulin instead lowers glucose ($r=-0.931$, $-0.904$).
This is consistent with treatment-confounding shortcut learning rather than a
model of insulin's mechanistic effect. The misleading association is larger
relative to the controlled response for basal: observational basal delivery
has slope $+65.5$~mg/dL/U versus $-11.5$ under perturbation; corresponding bolus
slopes are $+7.5$ and $-3.3$. This difference may help explain the worse basal
eRMSE, direction, and ranking. These associations do not establish a unique
cause or show that a given architecture cannot learn the correct mechanism.
Neither autoregression nor a low parameter count is sufficient protection
in these experiments.

\paragraph{Hypoglycemia Warning.}
Near-onset detection and advance warning are different requirements. Reporting
recall across minimum lead times at a stated alarm budget makes their trade-off
visible \citep{harvey2012clinically}. The benefit depends on the response:
suspending delivery cannot remove insulin already delivered, carbohydrate
absorption takes time, and caretaker response differs with living arrangements.
Repeated unnecessary alerts can reduce responsiveness \citep{shivers2013turn}.
Overnight evaluation is therefore particularly relevant.
IOBP2's frequent small automatic boluses give its post-bolus hypoglycemia
slice a different interpretation from the less frequent, larger boluses in
the other cohorts.

\paragraph{Scope and Limitations.}
Our insulin-response evaluation uses simulated rather than real-world
intervention data. We limit our comparison to point forecasts and fully
data-driven glucose predictors. Probabilistic forecasting can separate
trajectory prediction from risk-aware decision making \citep{namazi2025softtokens}.
Physiology-informed modeling is another promising direction
\citep{fathkouhi2026interventional,zou2024hybrid}.

\section{Conclusion}
We identify two weaknesses in promising uses of CGM forecasts: aggregate alarm
recall can hide differences in the advance warning available across event
contexts, and forecasts can reverse insulin effects or choose poor actions
despite better effect direction and ordering. Warning-time curves and paired
insulin perturbations make these weaknesses directly testable. We provide
evaluation code and a simulated intervention dataset to support model assessment
before prospective or closed-loop validation.

\label{sec:main-end}
\FloatBarrier
\acks{This work was partially supported by the LaunchPad for Diabetes program.}
\bibliography{references,camera_ready_additions}
\clearpage
\onecolumn
\raggedbottom
\appendix

\section{Cohort Descriptions and Preprocessing}
\label{app:data}
\footnotesize
\setlength{\tabcolsep}{4pt}
\begin{longtable}{@{}llcc@{}}
\caption{\textbf{Clinical cohort descriptions.} Glucose and per-patient event counts: median [Q1, Q3] over the processed cohort; events are follow-up totals, not daily rates.}\label{tab:cohort-details}\\
\toprule
Cohort & Train/validation/test & CGM (mg/dL) & Events per patient\\\midrule
\endfirsthead
\multicolumn{4}{c}{\tablename\ \thetable\ (continued)}\\\toprule
Cohort & Train/validation/test & CGM (mg/dL) & Events per patient\\\midrule
\endhead
\bottomrule\endfoot
DCLP3 & 79/11/24 & 147.0 [117.0, 189.0] & 53.0 [23.0, 91.0] \\
DCLP5 & 70/10/20 & 149.0 [117.0, 199.5] & 110.5 [60.8, 180.5] \\
PEDAP & 69/9/21 & 143.0 [112.0, 190.0] & 190.0 [116.5, 314.0] \\
IOBP2 & 230/33/67 & 150.0 [119.0, 191.0] & 46.0 [27.0, 72.0] \\
\end{longtable}
\normalsize

We retain numeric CGM readings with valid timestamps, deduplicate within
subjects, split records at prolonged gaps, and align each continuous segment
to a five-minute grid. For DCLP3, DCLP5, and PEDAP, the deduplication tolerance
is 15 seconds and the segment split threshold is 30 minutes. Glucose is
linearly interpolated within segments, never across segment boundaries.
Basal delivery is aligned as a piecewise-constant rate with forward filling;
missing initial rates are set to zero. Bolus records are aggregated in their
five-minute bins, with absent events set to zero. The alignment windows are
3 hours before/15 seconds after the grid time for basal records, and 285
seconds before/15 seconds after for bolus records. Meal information, when
present, is aligned as an event channel. IOBP2 records categorical meal
announcements. The processed IOBP2 meal channel is zero throughout.
These retrospective preprocessing choices do not
constitute a validated prospective missing-data strategy.

Records have subject and sequence identifiers, time, glucose, and available
basal, bolus, and meal channels. Optional weight is a slowly varying covariate.
Splits are assigned at the patient level before generating model windows;
IOB histories reset at sequence gaps. Table~\ref{tab:cohort-details} reports
glucose quartiles over pooled processed readings and hypoglycemic-event
quartiles across patients over their recorded follow-up. These descriptive
counts are not the eligible warning-curve denominators.
\FloatBarrier

\section{Alarm Algorithm and Calibration Details}
\label{app:alarm}
\begin{algorithm2e}[t]
\caption{Full-timeline alarm matching before event-context evaluation.}
\label{alg:alarm-matching}
\textbf{Input:} Processed CGM, forecast minima at eligible origins, validation
target $b$, and required warning times $\ell=0,5,\ldots,60$ minutes.
\begin{enumerate}
\item Extract low onsets separately in each continuous segment using three
low samples and three recovered samples. Exclude lows present at segment start.
\item On validation patients, scan alarm cutoffs. For each cutoff, keep the
first alarm and wait at least 60 minutes before keeping another. Match alarms
one-to-one to event onsets in the following 60 minutes, strictly before onset.
Use the full timeline for matching before assigning events to slices.
\item Select the cutoff with validation false alarms/day closest to but not
above $b$. Freeze it and repeat full-timeline alarm generation and matching on test.
\item Count unmatched retained alarms as false and divide by each patient's
full evaluable observation time.
\item Retain recall events with a complete preceding 60-minute eligible-origin
window. Assign contexts from event onset.
\item For each $\ell$, count retained events with a matched warning at least
$\ell$ minutes early, keeping events and matches fixed.
\item Macro-average over patients and bootstrap patients for pointwise intervals.
\end{enumerate}
\textbf{Output:} Test false alarms/day and event recall at each $\ell$.
\end{algorithm2e}

All hypoglycemia warning models are evaluated at the same prediction origins,
which require a complete processed 24-hour history and 60-minute target window.
Timestamp gaps restart event extraction. The matching pool
includes onsets with at least one preceding eligible origin; recall curves
additionally require all 12 preceding five-minute origins. The matching pool
is therefore not filtered by the requested warning time. This avoids changing
false-alarm classification or recall denominators when the requirement changes.
The 60-minute point credits alarms exactly 60 minutes before onset, because
earlier alarms lie outside the fixed matching window.

Calibration scans cutoffs in mg/dL and refines around feasible operating
points. False-alarm rate after refractory suppression need not vary monotonically
with cutoff. Ties in distance below the budget favor integrated validation
warning recall, then recall at zero additional warning, then the lower cutoff.
Neither test outcomes nor test-slice recall select thresholds.
Actual test rates and selected cutoffs are reported with the supplementary
warning results. No-event patients contribute to false-alarm estimates but
not to recall for the corresponding slice. Paired contrasts resample the same
event-bearing patients on both sides; they need not equal subtraction of the
separately macro-averaged slice means.
\FloatBarrier

\section{Forecaster Implementation Details}
\label{app:models}
We use no pretrained weights.

\paragraph{Simulator Linear Baseline.}
For glucose target $g_t$, the selected one-step model is
\begin{equation}
\hat g_t=c+\sum_{j=1}^{2}a_jg_{t-j}
+\sum_{q=0}^{47}\bigl(b_q u^{\mathrm{basal}}_{t-q}
+d_q u^{\mathrm{bolus}}_{t-q}\bigr).
\end{equation}
The insulin value at timestamp $t$ is an input for the glucose target at $t$;
glucose regressors use earlier timestamps only. In future rollout, insulin
at the predicted timestamp comes from the supplied plan. There are 98 input
coefficients and an intercept. Training uses all eligible factual training
rows, train-only standardization, an unpenalized intercept, and squared error
with ridge regularization. Search ranges are glucose lags
$\{2,3,6,12,24,48\}$, insulin inputs per channel $\{1,6,12,24,48\}$ including
the current timestamp, differencing $\{0,1\}$, and penalties
$\{0,10^{-4},10^{-2},1\}$. The selected model has validation rollout RMSE
27.37~mg/dL and autoregressive spectral radius 0.987. We evaluate all 240
configurations and select the lowest four-hour factual validation RMSE among
238 finite, stable fits. The selected specification has no moving-average
term or differencing and ridge penalty 0.01; it is fitted again on training
data before both intervention menus are evaluated. Selection uses pooled
four-hour factual validation trajectory error, not counterfactual metrics.

\paragraph{IOB Feature and Simple Forecaster.}
For dose $D$ delivered $a$ minutes earlier, the remaining-insulin kernel is
\begin{equation}
I(a;D)=D(1+a/\tau)\exp(-a/\tau),\qquad \tau=100\ \mathrm{min},
\end{equation}
following \citet{lee2009mpciob}, Eq.~(1). We sum recorded boluses and basal
delivery converted from U/hour to U per five-minute interval, over up to 24
hours within each sequence. This is modeled total IOB, not proprietary pump
displayed or net IOB. The feature vector contains latest glucose, backward
five-minute glucose derivative, and total IOB. For each lead $k$, ridge
regression predicts glucose change from that feature vector and adds it to
latest glucose. Scaling and coefficients use training patients; validation
patient-macro trajectory RMSE selects regularization. No alarm labels are
used to fit the forecaster. Its coefficient signs are unconstrained.

\paragraph{Neural Inputs.}
Clinical models use their selected history lengths and recorded channels.
The simulator models receive CGM, basal, and bolus. For AttenGluco, basal
and bolus occupy its separate auxiliary streams; its original three-stream
interface does not add a separate meal channel. The simulator future-plan
encoding shifts insulin inputs while leaving CGM history unchanged and is
identical in fitting and evaluation. Multi-step neural predictions are direct
trajectory outputs; only the simulator linear baseline uses recursive glucose
feedback. \glimmer{} changes PatchMLP's training loss, not its input architecture.
\FloatBarrier

\section{Additional Forecasting Results}
\label{app:forecasting}
Tables in this section report 30- and 60-minute point-forecast metrics with
95\% patient-bootstrap intervals. They retain their forecast-origin context
definitions, unlike the onset-defined alarm slices. No alarm recall, false-alarm,
or warning-time metric is inferred from these point-forecast tables.
\footnotesize
\setlength{\tabcolsep}{4pt}
\begin{longtable}{llcccc}
\caption{\textbf{Overall: point-forecast metrics.} RMSE (mg/dL) and PEG C--E\%, patient means [95\% CI]. These are not alarm metrics.}\label{tab:forecast-overall}\\
\toprule
Cohort & Forecaster & 30-min RMSE & 30-min PEG C--E\% & 60-min RMSE & 60-min PEG C--E\%\\\midrule
\endfirsthead
\multicolumn{6}{c}{\tablename\ \thetable\ (continued)}\\\toprule
Cohort & Forecaster & 30-min RMSE & 30-min PEG C--E\% & 60-min RMSE & 60-min PEG C--E\%\\\midrule
\endhead
\bottomrule\endfoot
DCLP3 & ARIMAX & 14.1 [13.1, 15.0] & 0.10 [0.07, 0.12] & 24.4 [22.8, 25.9] & 0.84 [0.69, 1.01] \\
DCLP3 & DLinear & 12.8 [12.0, 13.6] & 0.06 [0.05, 0.08] & 23.1 [21.6, 24.4] & 0.70 [0.58, 0.84] \\
DCLP3 & GRU & 12.2 [11.5, 13.0] & 0.09 [0.07, 0.11] & 22.3 [20.8, 23.6] & 0.69 [0.57, 0.85] \\
DCLP3 & iTransformer & 13.4 [12.5, 14.2] & 0.08 [0.06, 0.10] & 23.4 [21.9, 24.8] & 0.78 [0.66, 0.89] \\
DCLP3 & PatchMLP & 12.8 [12.0, 13.6] & 0.08 [0.06, 0.10] & 23.5 [22.1, 24.9] & 0.60 [0.50, 0.71] \\
DCLP3 & AttenGluco & 12.3 [11.5, 13.1] & 0.09 [0.07, 0.10] & 22.0 [20.6, 23.2] & 0.64 [0.55, 0.74] \\
DCLP3 & Glimmer\textsuperscript{*} & 13.6 [12.7, 14.4] & 0.04 [0.03, 0.05] & 22.8 [21.4, 24.2] & 0.67 [0.52, 0.89] \\
DCLP3 & ZOH & 17.3 [16.2, 18.2] & 0.15 [0.11, 0.18] & 27.9 [26.3, 29.3] & 0.96 [0.78, 1.13] \\
\end{longtable}
\normalsize
\footnotesize
\setlength{\tabcolsep}{4pt}
\begin{longtable}{llcccc}
\caption{\textbf{Post-bolus forecast origins: point-forecast metrics.} RMSE (mg/dL) and PEG C--E\%, patient means [95\% CI]. These are not alarm metrics.}\label{tab:forecast-post_bolus}\\
\toprule
Cohort & Forecaster & 30-min RMSE & 30-min PEG C--E\% & 60-min RMSE & 60-min PEG C--E\%\\\midrule
\endfirsthead
\multicolumn{6}{c}{\tablename\ \thetable\ (continued)}\\\toprule
Cohort & Forecaster & 30-min RMSE & 30-min PEG C--E\% & 60-min RMSE & 60-min PEG C--E\%\\\midrule
\endhead
\bottomrule\endfoot
DCLP3 & ARIMAX & 17.3 [16.1, 18.5] & 0.12 [0.08, 0.16] & 29.4 [27.6, 31.2] & 1.11 [0.87, 1.35] \\
DCLP3 & DLinear & 15.9 [14.7, 16.9] & 0.08 [0.05, 0.10] & 28.4 [26.5, 30.1] & 1.26 [0.87, 1.85] \\
DCLP3 & GRU & 15.2 [14.2, 16.2] & 0.10 [0.06, 0.16] & 27.7 [25.8, 29.7] & 1.09 [0.75, 1.59] \\
DCLP3 & iTransformer & 16.5 [15.4, 17.5] & 0.09 [0.06, 0.13] & 29.0 [27.1, 30.9] & 1.14 [0.90, 1.42] \\
DCLP3 & PatchMLP & 15.9 [14.8, 16.8] & 0.07 [0.05, 0.10] & 28.4 [26.7, 30.0] & 0.78 [0.65, 0.92] \\
DCLP3 & AttenGluco & 15.4 [14.4, 16.4] & 0.09 [0.05, 0.15] & 27.5 [25.6, 29.3] & 1.03 [0.75, 1.41] \\
DCLP3 & Glimmer\textsuperscript{*} & 16.6 [15.5, 17.6] & 0.06 [0.04, 0.08] & 28.6 [26.7, 30.5] & 1.25 [0.85, 1.87] \\
DCLP3 & ZOH & 21.7 [20.2, 23.1] & 0.20 [0.15, 0.25] & 34.7 [32.6, 36.6] & 1.49 [1.22, 1.79] \\
\end{longtable}
\normalsize
\footnotesize
\setlength{\tabcolsep}{4pt}
\begin{longtable}{llcccc}
\caption{\textbf{Nocturnal forecast origins: point-forecast metrics.} RMSE (mg/dL) and PEG C--E\%, patient means [95\% CI]. These are not alarm metrics.}\label{tab:forecast-nocturnal}\\
\toprule
Cohort & Forecaster & 30-min RMSE & 30-min PEG C--E\% & 60-min RMSE & 60-min PEG C--E\%\\\midrule
\endfirsthead
\multicolumn{6}{c}{\tablename\ \thetable\ (continued)}\\\toprule
Cohort & Forecaster & 30-min RMSE & 30-min PEG C--E\% & 60-min RMSE & 60-min PEG C--E\%\\\midrule
\endhead
\bottomrule\endfoot
DCLP3 & ARIMAX & 10.4 [9.7, 11.1] & 0.06 [0.03, 0.10] & 17.4 [16.4, 18.5] & 0.40 [0.28, 0.54] \\
DCLP3 & DLinear & 9.7 [8.9, 10.6] & 0.05 [0.02, 0.09] & 16.2 [15.1, 17.3] & 0.27 [0.17, 0.37] \\
DCLP3 & GRU & 9.3 [8.5, 10.1] & 0.08 [0.05, 0.12] & 15.9 [14.8, 17.1] & 0.34 [0.22, 0.48] \\
DCLP3 & iTransformer & 10.1 [9.3, 11.0] & 0.06 [0.03, 0.10] & 16.6 [15.4, 17.9] & 0.39 [0.25, 0.55] \\
DCLP3 & PatchMLP & 9.8 [8.9, 10.6] & 0.06 [0.03, 0.10] & 16.7 [15.6, 17.9] & 0.31 [0.20, 0.43] \\
DCLP3 & AttenGluco & 9.2 [8.4, 10.1] & 0.07 [0.04, 0.11] & 15.1 [14.0, 16.2] & 0.26 [0.17, 0.35] \\
DCLP3 & Glimmer\textsuperscript{*} & 10.3 [9.4, 11.2] & 0.03 [0.01, 0.06] & 15.6 [14.4, 16.8] & 0.22 [0.14, 0.32] \\
DCLP3 & ZOH & 11.6 [10.8, 12.6] & 0.06 [0.03, 0.09] & 18.3 [17.1, 19.7] & 0.28 [0.19, 0.38] \\
\end{longtable}
\normalsize

\FloatBarrier

\section{Warning-Time Results and Event Support}
\label{app:warning-results}
\footnotesize
\setlength{\tabcolsep}{4pt}
\begin{longtable}{llrrrr}
\caption{\textbf{Eligible test-event support.} Observation days cover the full evaluable timeline and are shared across slices and forecasters.}\label{tab:warning-support}\\
\toprule
Cohort & Event context & Patients & With events & Events & Observation days\\\midrule
\endfirsthead
\multicolumn{6}{c}{\tablename\ \thetable\ (continued)}\\\toprule
Cohort & Event context & Patients & With events & Events & Observation days\\\midrule
\endhead
\bottomrule\endfoot
DCLP3 & Overall & 24 & 23 & 641 & 1516.2 \\
DCLP3 & Post-bolus & 24 & 11 & 32 & 1516.2 \\
DCLP3 & Non-post-bolus & 24 & 23 & 609 & 1516.2 \\
DCLP3 & Nocturnal & 24 & 22 & 102 & 1516.2 \\
DCLP3 & Daytime & 24 & 22 & 539 & 1516.2 \\
\midrule
DCLP5 & Overall & 20 & 20 & 1967 & 2041.0 \\
DCLP5 & Post-bolus & 20 & 20 & 305 & 2041.0 \\
DCLP5 & Non-post-bolus & 20 & 20 & 1662 & 2041.0 \\
DCLP5 & Nocturnal & 20 & 20 & 250 & 2041.0 \\
DCLP5 & Daytime & 20 & 20 & 1717 & 2041.0 \\
\midrule
PEDAP & Overall & 21 & 21 & 2786 & 2773.0 \\
PEDAP & Post-bolus & 21 & 21 & 424 & 2773.0 \\
PEDAP & Non-post-bolus & 21 & 21 & 2362 & 2773.0 \\
PEDAP & Nocturnal & 21 & 21 & 467 & 2773.0 \\
PEDAP & Daytime & 21 & 21 & 2319 & 2773.0 \\
PEDAP & Post-meal (15 min) & 21 & 17 & 145 & 2773.0 \\
PEDAP & Other meal contexts & 21 & 21 & 2641 & 2773.0 \\
\midrule
IOBP2 & Overall & 67 & 67 & 2205 & 3206.0 \\
IOBP2 & Post-bolus & 67 & 48 & 113 & 3206.0 \\
IOBP2 & Non-post-bolus & 67 & 67 & 2092 & 3206.0 \\
IOBP2 & Nocturnal & 67 & 65 & 434 & 3206.0 \\
IOBP2 & Daytime & 67 & 67 & 1771 & 3206.0 \\
\midrule
\end{longtable}
\normalsize

The main setting uses a 2/day validation target and 15-minute minimum warning.
The following tables provide all models, including IOB-linear and persistence,
other event slices, warning requirements, and validation targets. Recall
intervals are percentages; differences are percentage points. Actual test
false-alarm rates are full-timeline rates, shared across event slices and
warning requirements. All curves show every five-minute requirement.
\footnotesize
\setlength{\tabcolsep}{4pt}
\begin{longtable}{llcccc}
\caption{\textbf{Warning recall: 0/5 minutes minimum, 2 FA/day validation target.} Test FA/day and recall (\%) [95\% CI]. Cutoffs are fixed across contexts and warning requirements.}\label{tab:warning-b2-l0}\\
\toprule
Cohort & Forecaster & Test FA/day & Overall & Post-bolus & Non-post-bolus\\\midrule
\endfirsthead
\multicolumn{6}{c}{\tablename\ \thetable\ (continued)}\\\toprule
Cohort & Forecaster & Test FA/day & Overall & Post-bolus & Non-post-bolus\\\midrule
\endhead
\bottomrule\endfoot
DCLP3 ($n=24$) & ARIMAX & 2.51 [1.85, 3.50] & 99.8 [99.4, 100.0] & 100.0 [100.0, 100.0] & 99.8 [99.4, 100.0] \\
 & DLinear & 2.56 [1.89, 3.54] & 99.8 [99.4, 100.0] & 100.0 [100.0, 100.0] & 99.8 [99.4, 100.0] \\
 & GRU & 2.78 [1.98, 3.98] & 99.8 [99.4, 100.0] & 100.0 [100.0, 100.0] & 99.8 [99.4, 100.0] \\
 & iTransformer & 2.74 [1.84, 4.17] & 97.4 [95.5, 99.1] & 97.0 [90.9, 100.0] & 97.5 [95.6, 99.2] \\
 & PatchMLP & 2.54 [1.86, 3.53] & 99.9 [99.7, 100.0] & 100.0 [100.0, 100.0] & 99.9 [99.7, 100.0] \\
 & AttenGluco & 2.69 [1.91, 3.88] & 99.8 [99.4, 100.0] & 100.0 [100.0, 100.0] & 99.8 [99.4, 100.0] \\
 & Glimmer\textsuperscript{*} & 2.60 [1.93, 3.56] & 99.9 [99.7, 100.0] & 100.0 [100.0, 100.0] & 99.9 [99.7, 100.0] \\
 & IOB-linear & 2.66 [1.91, 3.85] & 99.8 [99.4, 100.0] & 100.0 [100.0, 100.0] & 99.8 [99.4, 100.0] \\
\end{longtable}
\normalsize
\footnotesize
\setlength{\tabcolsep}{4pt}
\begin{longtable}{llcccc}
\caption{\textbf{Warning recall: 10 minutes minimum, 2 FA/day validation target.} Test FA/day and recall (\%) [95\% CI]. Cutoffs are fixed across contexts and warning requirements.}\label{tab:warning-b2-l10}\\
\toprule
Cohort & Forecaster & Test FA/day & Overall & Post-bolus & Non-post-bolus\\\midrule
\endfirsthead
\multicolumn{6}{c}{\tablename\ \thetable\ (continued)}\\\toprule
Cohort & Forecaster & Test FA/day & Overall & Post-bolus & Non-post-bolus\\\midrule
\endhead
\bottomrule\endfoot
DCLP3 ($n=24$) & ARIMAX & 2.51 [1.85, 3.50] & 97.1 [95.8, 98.4] & 98.7 [96.1, 100.0] & 97.0 [95.6, 98.4] \\
 & DLinear & 2.56 [1.89, 3.54] & 97.8 [96.5, 99.0] & 98.7 [96.1, 100.0] & 97.7 [96.3, 99.0] \\
 & GRU & 2.78 [1.98, 3.98] & 96.7 [94.6, 98.5] & 97.7 [93.2, 100.0] & 96.7 [94.6, 98.5] \\
 & iTransformer & 2.74 [1.84, 4.17] & 90.5 [85.9, 94.5] & 79.2 [51.9, 100.0] & 90.9 [86.3, 94.8] \\
 & PatchMLP & 2.54 [1.86, 3.53] & 98.4 [97.0, 99.6] & 98.7 [96.1, 100.0] & 98.4 [97.0, 99.6] \\
 & AttenGluco & 2.69 [1.91, 3.88] & 96.9 [95.0, 98.4] & 95.7 [89.6, 100.0] & 96.9 [95.1, 98.4] \\
 & Glimmer\textsuperscript{*} & 2.60 [1.93, 3.56] & 97.3 [95.7, 98.7] & 98.7 [96.1, 100.0] & 97.2 [95.6, 98.6] \\
 & IOB-linear & 2.66 [1.91, 3.85] & 97.7 [96.4, 98.9] & 98.7 [96.1, 100.0] & 97.6 [96.4, 98.8] \\
\end{longtable}
\normalsize
\footnotesize
\setlength{\tabcolsep}{4pt}
\begin{longtable}{llcccc}
\caption{\textbf{Warning recall: 15 minutes minimum, 2 FA/day validation target.} Test FA/day and recall (\%) [95\% CI]. Cutoffs are fixed across contexts and warning requirements.}\label{tab:warning-b2-l15}\\
\toprule
Cohort & Forecaster & Test FA/day & Overall & Post-bolus & Non-post-bolus\\\midrule
\endfirsthead
\multicolumn{6}{c}{\tablename\ \thetable\ (continued)}\\\toprule
Cohort & Forecaster & Test FA/day & Overall & Post-bolus & Non-post-bolus\\\midrule
\endhead
\bottomrule\endfoot
DCLP3 ($n=24$) & ARIMAX & 2.51 [1.85, 3.50] & 84.9 [80.8, 89.0] & 63.2 [40.6, 83.6] & 85.6 [81.6, 89.5] \\
 & DLinear & 2.56 [1.89, 3.54] & 91.6 [88.9, 94.2] & 86.6 [66.6, 100.0] & 91.5 [88.7, 94.1] \\
 & GRU & 2.78 [1.98, 3.98] & 88.6 [84.9, 91.8] & 69.4 [47.0, 89.5] & 89.2 [85.6, 92.3] \\
 & iTransformer & 2.74 [1.84, 4.17] & 78.4 [72.5, 84.0] & 62.5 [36.4, 85.5] & 79.0 [73.1, 84.6] \\
 & PatchMLP & 2.54 [1.86, 3.53] & 87.1 [82.9, 90.9] & 73.2 [48.5, 93.9] & 87.4 [83.2, 91.1] \\
 & AttenGluco & 2.69 [1.91, 3.88] & 88.8 [85.9, 91.9] & 65.6 [41.8, 86.4] & 89.3 [86.6, 92.3] \\
 & Glimmer\textsuperscript{*} & 2.60 [1.93, 3.56] & 89.7 [86.7, 92.8] & 71.6 [47.4, 91.8] & 90.2 [87.3, 93.2] \\
 & IOB-linear & 2.66 [1.91, 3.85] & 90.8 [88.3, 93.3] & 82.0 [62.1, 97.4] & 91.0 [88.6, 93.6] \\
\end{longtable}
\normalsize
\footnotesize
\setlength{\tabcolsep}{4pt}
\begin{longtable}{llcccc}
\caption{\textbf{Warning recall: 20 minutes minimum, 2 FA/day validation target.} Test FA/day and recall (\%) [95\% CI]. Cutoffs are fixed across contexts and warning requirements.}\label{tab:warning-b2-l20}\\
\toprule
Cohort & Forecaster & Test FA/day & Overall & Post-bolus & Non-post-bolus\\\midrule
\endfirsthead
\multicolumn{6}{c}{\tablename\ \thetable\ (continued)}\\\toprule
Cohort & Forecaster & Test FA/day & Overall & Post-bolus & Non-post-bolus\\\midrule
\endhead
\bottomrule\endfoot
DCLP3 ($n=24$) & ARIMAX & 2.51 [1.85, 3.50] & 70.2 [64.6, 76.2] & 58.9 [33.9, 81.6] & 70.6 [65.1, 76.7] \\
 & DLinear & 2.56 [1.89, 3.54] & 76.0 [70.9, 81.0] & 64.2 [40.3, 85.7] & 76.5 [71.4, 81.5] \\
 & GRU & 2.78 [1.98, 3.98] & 75.1 [69.1, 81.0] & 57.9 [34.1, 79.0] & 75.7 [69.7, 81.5] \\
 & iTransformer & 2.74 [1.84, 4.17] & 65.9 [59.0, 72.8] & 46.8 [23.0, 70.0] & 66.5 [59.5, 73.2] \\
 & PatchMLP & 2.54 [1.86, 3.53] & 73.7 [68.3, 79.4] & 46.8 [22.4, 69.8] & 74.3 [68.9, 79.9] \\
 & AttenGluco & 2.69 [1.91, 3.88] & 75.3 [70.0, 80.9] & 54.4 [29.1, 79.5] & 76.0 [70.8, 81.4] \\
 & Glimmer\textsuperscript{*} & 2.60 [1.93, 3.56] & 77.5 [72.7, 82.6] & 58.9 [34.1, 81.6] & 78.1 [73.3, 83.3] \\
 & IOB-linear & 2.66 [1.91, 3.85] & 78.5 [75.2, 81.8] & 67.2 [43.9, 87.9] & 78.8 [75.6, 82.3] \\
\end{longtable}
\normalsize
\footnotesize
\setlength{\tabcolsep}{4pt}
\begin{longtable}{llcccc}
\caption{\textbf{Warning recall: 25 minutes minimum, 2 FA/day validation target.} Test FA/day and recall (\%) [95\% CI]. Cutoffs are fixed across contexts and warning requirements.}\label{tab:warning-b2-l25}\\
\toprule
Cohort & Forecaster & Test FA/day & Overall & Post-bolus & Non-post-bolus\\\midrule
\endfirsthead
\multicolumn{6}{c}{\tablename\ \thetable\ (continued)}\\\toprule
Cohort & Forecaster & Test FA/day & Overall & Post-bolus & Non-post-bolus\\\midrule
\endhead
\bottomrule\endfoot
DCLP3 ($n=24$) & ARIMAX & 2.51 [1.85, 3.50] & 54.2 [47.1, 61.4] & 40.9 [20.0, 61.8] & 54.6 [47.6, 61.7] \\
 & DLinear & 2.56 [1.89, 3.54] & 61.6 [55.8, 67.9] & 46.2 [21.4, 68.6] & 62.1 [56.2, 68.4] \\
 & GRU & 2.78 [1.98, 3.98] & 61.8 [56.3, 67.9] & 45.3 [22.1, 67.1] & 62.2 [56.7, 68.2] \\
 & iTransformer & 2.74 [1.84, 4.17] & 52.6 [44.4, 61.2] & 45.5 [21.5, 68.8] & 52.9 [44.6, 61.5] \\
 & PatchMLP & 2.54 [1.86, 3.53] & 58.4 [50.5, 66.5] & 40.9 [20.0, 61.8] & 58.9 [51.1, 67.1] \\
 & AttenGluco & 2.69 [1.91, 3.88] & 64.0 [57.7, 70.7] & 41.7 [18.1, 66.0] & 64.5 [58.5, 71.2] \\
 & Glimmer\textsuperscript{*} & 2.60 [1.93, 3.56] & 63.9 [57.3, 70.8] & 43.2 [20.7, 65.0] & 64.4 [57.8, 71.2] \\
 & IOB-linear & 2.66 [1.91, 3.85] & 62.0 [56.2, 68.3] & 46.2 [21.4, 68.6] & 62.5 [56.7, 68.8] \\
\end{longtable}
\normalsize
\footnotesize
\setlength{\tabcolsep}{4pt}
\begin{longtable}{llcccc}
\caption{\textbf{Warning recall: 30 minutes minimum, 2 FA/day validation target.} Test FA/day and recall (\%) [95\% CI]. Cutoffs are fixed across contexts and warning requirements.}\label{tab:warning-b2-l30}\\
\toprule
Cohort & Forecaster & Test FA/day & Overall & Post-bolus & Non-post-bolus\\\midrule
\endfirsthead
\multicolumn{6}{c}{\tablename\ \thetable\ (continued)}\\\toprule
Cohort & Forecaster & Test FA/day & Overall & Post-bolus & Non-post-bolus\\\midrule
\endhead
\bottomrule\endfoot
DCLP3 ($n=24$) & ARIMAX & 2.51 [1.85, 3.50] & 42.5 [35.5, 50.6] & 32.5 [13.0, 53.9] & 42.8 [35.8, 50.8] \\
 & DLinear & 2.56 [1.89, 3.54] & 44.8 [37.2, 53.2] & 38.4 [17.5, 59.7] & 45.0 [37.5, 53.4] \\
 & GRU & 2.78 [1.98, 3.98] & 48.3 [41.6, 55.7] & 42.1 [19.7, 63.6] & 48.5 [41.9, 55.8] \\
 & iTransformer & 2.74 [1.84, 4.17] & 43.6 [35.8, 52.3] & 43.2 [20.8, 66.4] & 43.6 [35.9, 52.1] \\
 & PatchMLP & 2.54 [1.86, 3.53] & 43.3 [36.2, 51.1] & 36.1 [17.0, 56.5] & 43.5 [36.5, 51.5] \\
 & AttenGluco & 2.69 [1.91, 3.88] & 50.9 [43.6, 58.4] & 38.1 [16.2, 60.0] & 51.2 [44.0, 58.6] \\
 & Glimmer\textsuperscript{*} & 2.60 [1.93, 3.56] & 52.6 [46.8, 59.3] & 43.2 [20.7, 65.0] & 52.7 [47.1, 59.2] \\
 & IOB-linear & 2.66 [1.91, 3.85] & 45.5 [38.1, 53.5] & 38.4 [17.5, 59.7] & 45.6 [38.3, 53.6] \\
\end{longtable}
\normalsize
\footnotesize
\setlength{\tabcolsep}{4pt}
\begin{longtable}{llcccc}
\caption{\textbf{Warning recall: 35 minutes minimum, 2 FA/day validation target.} Test FA/day and recall (\%) [95\% CI]. Cutoffs are fixed across contexts and warning requirements.}\label{tab:warning-b2-l35}\\
\toprule
Cohort & Forecaster & Test FA/day & Overall & Post-bolus & Non-post-bolus\\\midrule
\endfirsthead
\multicolumn{6}{c}{\tablename\ \thetable\ (continued)}\\\toprule
Cohort & Forecaster & Test FA/day & Overall & Post-bolus & Non-post-bolus\\\midrule
\endhead
\bottomrule\endfoot
DCLP3 ($n=24$) & ARIMAX & 2.51 [1.85, 3.50] & 33.6 [26.5, 42.1] & 30.2 [12.6, 50.0] & 33.7 [26.5, 42.2] \\
 & DLinear & 2.56 [1.89, 3.54] & 35.6 [28.2, 44.2] & 38.4 [17.5, 59.7] & 35.3 [28.0, 44.1] \\
 & GRU & 2.78 [1.98, 3.98] & 38.7 [31.8, 46.9] & 36.8 [17.7, 55.9] & 38.7 [31.8, 46.7] \\
 & iTransformer & 2.74 [1.84, 4.17] & 35.2 [28.3, 43.6] & 36.8 [17.5, 57.0] & 35.1 [28.3, 43.4] \\
 & PatchMLP & 2.54 [1.86, 3.53] & 36.0 [29.0, 44.3] & 36.1 [17.0, 56.5] & 35.9 [28.9, 44.3] \\
 & AttenGluco & 2.69 [1.91, 3.88] & 40.4 [33.4, 48.2] & 31.0 [11.5, 51.8] & 40.6 [33.6, 48.3] \\
 & Glimmer\textsuperscript{*} & 2.60 [1.93, 3.56] & 43.0 [36.5, 50.8] & 39.1 [18.4, 59.4] & 43.0 [36.4, 50.7] \\
 & IOB-linear & 2.66 [1.91, 3.85] & 36.3 [29.2, 44.8] & 36.1 [16.2, 56.4] & 36.2 [29.1, 44.6] \\
\end{longtable}
\normalsize
\footnotesize
\setlength{\tabcolsep}{4pt}
\begin{longtable}{llcccc}
\caption{\textbf{Warning recall: 40 minutes minimum, 2 FA/day validation target.} Test FA/day and recall (\%) [95\% CI]. Cutoffs are fixed across contexts and warning requirements.}\label{tab:warning-b2-l40}\\
\toprule
Cohort & Forecaster & Test FA/day & Overall & Post-bolus & Non-post-bolus\\\midrule
\endfirsthead
\multicolumn{6}{c}{\tablename\ \thetable\ (continued)}\\\toprule
Cohort & Forecaster & Test FA/day & Overall & Post-bolus & Non-post-bolus\\\midrule
\endhead
\bottomrule\endfoot
DCLP3 ($n=24$) & ARIMAX & 2.51 [1.85, 3.50] & 24.6 [17.5, 33.5] & 13.5 [3.0, 24.7] & 24.9 [17.8, 33.7] \\
 & DLinear & 2.56 [1.89, 3.54] & 27.1 [20.3, 36.3] & 32.0 [13.5, 52.4] & 26.9 [20.2, 35.9] \\
 & GRU & 2.78 [1.98, 3.98] & 30.8 [23.8, 39.4] & 30.7 [13.0, 49.8] & 30.7 [23.6, 39.4] \\
 & iTransformer & 2.74 [1.84, 4.17] & 28.7 [21.6, 37.3] & 30.2 [12.6, 50.0] & 28.6 [21.6, 37.1] \\
 & PatchMLP & 2.54 [1.86, 3.53] & 25.2 [18.3, 34.1] & 27.4 [12.1, 46.0] & 25.2 [18.3, 34.1] \\
 & AttenGluco & 2.69 [1.91, 3.88] & 33.9 [27.1, 42.0] & 25.2 [9.1, 44.9] & 34.3 [27.5, 42.5] \\
 & Glimmer\textsuperscript{*} & 2.60 [1.93, 3.56] & 34.4 [28.5, 42.1] & 34.6 [15.8, 53.4] & 34.2 [28.5, 41.9] \\
 & IOB-linear & 2.66 [1.91, 3.85] & 27.1 [19.9, 35.8] & 22.9 [8.9, 37.9] & 27.0 [19.8, 35.7] \\
\end{longtable}
\normalsize
\footnotesize
\setlength{\tabcolsep}{4pt}
\begin{longtable}{llcccc}
\caption{\textbf{Warning recall: 45 minutes minimum, 2 FA/day validation target.} Test FA/day and recall (\%) [95\% CI]. Cutoffs are fixed across contexts and warning requirements.}\label{tab:warning-b2-l45}\\
\toprule
Cohort & Forecaster & Test FA/day & Overall & Post-bolus & Non-post-bolus\\\midrule
\endfirsthead
\multicolumn{6}{c}{\tablename\ \thetable\ (continued)}\\\toprule
Cohort & Forecaster & Test FA/day & Overall & Post-bolus & Non-post-bolus\\\midrule
\endhead
\bottomrule\endfoot
DCLP3 ($n=24$) & ARIMAX & 2.51 [1.85, 3.50] & 16.9 [12.1, 22.2] & 11.7 [2.3, 22.0] & 17.1 [12.3, 22.4] \\
 & DLinear & 2.56 [1.89, 3.54] & 20.7 [13.8, 30.0] & 18.8 [5.6, 34.4] & 20.5 [13.3, 29.6] \\
 & GRU & 2.78 [1.98, 3.98] & 22.3 [15.4, 31.2] & 19.8 [7.1, 32.2] & 22.1 [15.1, 30.8] \\
 & iTransformer & 2.74 [1.84, 4.17] & 21.6 [15.4, 30.4] & 17.1 [6.8, 27.9] & 21.6 [15.4, 30.3] \\
 & PatchMLP & 2.54 [1.86, 3.53] & 18.0 [11.2, 27.1] & 14.3 [4.4, 24.9] & 18.1 [11.2, 27.1] \\
 & AttenGluco & 2.69 [1.91, 3.88] & 24.2 [17.3, 33.2] & 23.4 [7.6, 43.1] & 24.4 [17.5, 33.4] \\
 & Glimmer\textsuperscript{*} & 2.60 [1.93, 3.56] & 25.8 [19.3, 34.0] & 23.7 [9.1, 38.1] & 25.5 [18.9, 33.7] \\
 & IOB-linear & 2.66 [1.91, 3.85] & 21.5 [14.3, 30.4] & 18.8 [5.6, 34.4] & 21.3 [14.4, 30.4] \\
\end{longtable}
\normalsize
\footnotesize
\setlength{\tabcolsep}{4pt}
\begin{longtable}{llcccc}
\caption{\textbf{Warning recall: 50 minutes minimum, 2 FA/day validation target.} Test FA/day and recall (\%) [95\% CI]. Cutoffs are fixed across contexts and warning requirements.}\label{tab:warning-b2-l50}\\
\toprule
Cohort & Forecaster & Test FA/day & Overall & Post-bolus & Non-post-bolus\\\midrule
\endfirsthead
\multicolumn{6}{c}{\tablename\ \thetable\ (continued)}\\\toprule
Cohort & Forecaster & Test FA/day & Overall & Post-bolus & Non-post-bolus\\\midrule
\endhead
\bottomrule\endfoot
DCLP3 ($n=24$) & ARIMAX & 2.51 [1.85, 3.50] & 9.2 [6.1, 12.3] & 9.8 [0.0, 21.2] & 9.2 [6.1, 12.2] \\
 & DLinear & 2.56 [1.89, 3.54] & 13.3 [9.2, 17.7] & 15.7 [2.6, 32.4] & 13.1 [9.1, 17.7] \\
 & GRU & 2.78 [1.98, 3.98] & 13.9 [9.2, 19.1] & 13.4 [2.6, 26.8] & 13.8 [9.2, 19.0] \\
 & iTransformer & 2.74 [1.84, 4.17] & 13.2 [9.4, 17.7] & 7.1 [0.0, 14.2] & 13.6 [9.7, 18.4] \\
 & PatchMLP & 2.54 [1.86, 3.53] & 9.1 [6.0, 12.7] & 9.7 [2.3, 17.6] & 9.4 [6.0, 13.3] \\
 & AttenGluco & 2.69 [1.91, 3.88] & 16.1 [9.4, 25.4] & 5.8 [0.0, 11.7] & 16.5 [9.6, 25.9] \\
 & Glimmer\textsuperscript{*} & 2.60 [1.93, 3.56] & 19.6 [13.3, 28.2] & 18.8 [7.1, 30.6] & 19.2 [13.0, 27.9] \\
 & IOB-linear & 2.66 [1.91, 3.85] & 14.2 [10.0, 18.8] & 15.7 [2.6, 32.4] & 14.1 [9.9, 18.6] \\
\end{longtable}
\normalsize
\footnotesize
\setlength{\tabcolsep}{4pt}
\begin{longtable}{llcccc}
\caption{\textbf{Warning recall: 55 minutes minimum, 2 FA/day validation target.} Test FA/day and recall (\%) [95\% CI]. Cutoffs are fixed across contexts and warning requirements.}\label{tab:warning-b2-l55}\\
\toprule
Cohort & Forecaster & Test FA/day & Overall & Post-bolus & Non-post-bolus\\\midrule
\endfirsthead
\multicolumn{6}{c}{\tablename\ \thetable\ (continued)}\\\toprule
Cohort & Forecaster & Test FA/day & Overall & Post-bolus & Non-post-bolus\\\midrule
\endhead
\bottomrule\endfoot
DCLP3 ($n=24$) & ARIMAX & 2.51 [1.85, 3.50] & 5.6 [3.5, 7.9] & 0.0 [0.0, 0.0] & 6.0 [3.7, 8.4] \\
 & DLinear & 2.56 [1.89, 3.54] & 6.3 [3.7, 9.3] & 11.1 [0.0, 24.8] & 6.2 [3.5, 9.5] \\
 & GRU & 2.78 [1.98, 3.98] & 7.5 [5.1, 10.1] & 8.9 [0.0, 19.1] & 7.5 [5.1, 10.1] \\
 & iTransformer & 2.74 [1.84, 4.17] & 8.7 [5.0, 13.7] & 7.1 [0.0, 14.2] & 8.8 [5.1, 13.9] \\
 & PatchMLP & 2.54 [1.86, 3.53] & 5.6 [3.2, 8.7] & 2.6 [0.0, 7.8] & 5.9 [3.3, 9.4] \\
 & AttenGluco & 2.69 [1.91, 3.88] & 10.7 [4.6, 19.8] & 4.5 [0.0, 11.4] & 10.9 [4.8, 20.1] \\
 & Glimmer\textsuperscript{*} & 2.60 [1.93, 3.56] & 10.0 [7.2, 13.1] & 12.0 [2.6, 22.2] & 9.8 [6.8, 12.8] \\
 & IOB-linear & 2.66 [1.91, 3.85] & 6.0 [3.3, 9.0] & 5.8 [0.0, 14.9] & 6.1 [3.3, 9.1] \\
\end{longtable}
\normalsize
\footnotesize
\setlength{\tabcolsep}{4pt}
\begin{longtable}{llcccc}
\caption{\textbf{Warning recall: 60 minutes minimum, 2 FA/day validation target.} Test FA/day and recall (\%) [95\% CI]. Cutoffs are fixed across contexts and warning requirements.}\label{tab:warning-b2-l60}\\
\toprule
Cohort & Forecaster & Test FA/day & Overall & Post-bolus & Non-post-bolus\\\midrule
\endfirsthead
\multicolumn{6}{c}{\tablename\ \thetable\ (continued)}\\\toprule
Cohort & Forecaster & Test FA/day & Overall & Post-bolus & Non-post-bolus\\\midrule
\endhead
\bottomrule\endfoot
DCLP3 ($n=24$) & ARIMAX & 2.51 [1.85, 3.50] & 2.3 [1.0, 3.9] & 0.0 [0.0, 0.0] & 2.5 [1.1, 4.1] \\
 & DLinear & 2.56 [1.89, 3.54] & 3.3 [1.8, 5.2] & 2.3 [0.0, 6.8] & 3.4 [1.7, 5.4] \\
 & GRU & 2.78 [1.98, 3.98] & 3.8 [2.2, 5.7] & 3.0 [0.0, 9.1] & 3.9 [2.2, 6.0] \\
 & iTransformer & 2.74 [1.84, 4.17] & 3.0 [1.6, 4.6] & 4.8 [0.0, 11.5] & 3.0 [1.6, 4.7] \\
 & PatchMLP & 2.54 [1.86, 3.53] & 3.1 [1.6, 4.9] & 1.3 [0.0, 3.9] & 3.2 [1.6, 5.2] \\
 & AttenGluco & 2.69 [1.91, 3.88] & 2.8 [1.2, 4.6] & 0.0 [0.0, 0.0] & 3.0 [1.3, 5.0] \\
 & Glimmer\textsuperscript{*} & 2.60 [1.93, 3.56] & 4.1 [1.9, 7.0] & 4.3 [0.0, 10.4] & 4.1 [1.9, 6.9] \\
 & IOB-linear & 2.66 [1.91, 3.85] & 3.5 [1.7, 5.5] & 2.3 [0.0, 6.8] & 3.6 [1.7, 5.8] \\
\end{longtable}
\normalsize
\footnotesize
\setlength{\tabcolsep}{4pt}
\begin{longtable}{llcccc}
\caption{\textbf{Warning recall: 15 minutes minimum, 1 FA/day validation target.} Test FA/day and recall (\%) [95\% CI]. Cutoffs are fixed across contexts and warning requirements.}\label{tab:warning-b1-l15}\\
\toprule
Cohort & Forecaster & Test FA/day & Overall & Post-bolus & Non-post-bolus\\\midrule
\endfirsthead
\multicolumn{6}{c}{\tablename\ \thetable\ (continued)}\\\toprule
Cohort & Forecaster & Test FA/day & Overall & Post-bolus & Non-post-bolus\\\midrule
\endhead
\bottomrule\endfoot
DCLP3 ($n=24$) & ARIMAX & 1.31 [0.91, 1.92] & 63.3 [57.7, 68.7] & 46.7 [24.2, 70.0] & 64.2 [58.5, 69.5] \\
 & DLinear & 1.39 [0.97, 1.99] & 78.2 [72.3, 83.9] & 61.6 [37.9, 84.0] & 79.3 [73.7, 84.6] \\
 & GRU & 1.43 [0.98, 2.04] & 72.7 [63.9, 80.4] & 59.9 [34.4, 83.4] & 73.3 [64.5, 80.8] \\
 & iTransformer & 1.50 [0.97, 2.32] & 60.2 [51.4, 69.0] & 56.6 [36.4, 75.5] & 60.5 [51.7, 69.5] \\
 & PatchMLP & 1.30 [0.90, 1.90] & 63.4 [57.4, 69.7] & 39.9 [18.5, 60.6] & 64.3 [58.0, 70.7] \\
 & AttenGluco & 1.33 [0.91, 1.94] & 70.6 [64.4, 76.5] & 54.4 [28.9, 78.7] & 71.2 [65.2, 77.1] \\
 & Glimmer\textsuperscript{*} & 1.29 [0.98, 1.68] & 76.5 [71.8, 81.6] & 59.3 [35.8, 81.3] & 77.1 [72.5, 82.0] \\
 & IOB-linear & 1.38 [0.97, 1.97] & 76.6 [72.3, 81.0] & 66.9 [41.6, 87.9] & 77.3 [73.3, 81.6] \\
\end{longtable}
\normalsize
\footnotesize
\setlength{\tabcolsep}{4pt}
\begin{longtable}{llcccc}
\caption{\textbf{Warning recall: 15 minutes minimum, 1.5 FA/day validation target.} Test FA/day and recall (\%) [95\% CI]. Cutoffs are fixed across contexts and warning requirements.}\label{tab:warning-b1p5-l15}\\
\toprule
Cohort & Forecaster & Test FA/day & Overall & Post-bolus & Non-post-bolus\\\midrule
\endfirsthead
\multicolumn{6}{c}{\tablename\ \thetable\ (continued)}\\\toprule
Cohort & Forecaster & Test FA/day & Overall & Post-bolus & Non-post-bolus\\\midrule
\endhead
\bottomrule\endfoot
DCLP3 ($n=24$) & ARIMAX & 2.02 [1.39, 3.02] & 79.2 [73.9, 84.4] & 65.5 [41.9, 86.5] & 79.6 [74.3, 84.8] \\
 & DLinear & 2.04 [1.41, 3.04] & 87.7 [83.3, 91.9] & 80.7 [60.0, 95.5] & 88.0 [83.5, 92.2] \\
 & GRU & 2.05 [1.47, 2.85] & 84.4 [78.6, 89.7] & 68.6 [43.9, 90.3] & 84.9 [79.1, 90.3] \\
 & iTransformer & 2.05 [1.38, 3.07] & 72.7 [64.6, 80.4] & 67.2 [43.0, 87.8] & 73.2 [65.1, 80.7] \\
 & PatchMLP & 1.97 [1.32, 2.97] & 77.9 [72.2, 83.7] & 62.0 [36.4, 84.7] & 78.5 [72.9, 84.2] \\
 & AttenGluco & 2.02 [1.37, 3.02] & 84.5 [80.9, 88.4] & 65.0 [41.2, 86.2] & 85.0 [81.5, 88.8] \\
 & Glimmer\textsuperscript{*} & 2.02 [1.39, 3.02] & 86.0 [81.7, 90.3] & 65.0 [41.7, 86.4] & 86.8 [82.5, 91.0] \\
 & IOB-linear & 2.08 [1.45, 3.07] & 87.3 [83.6, 90.9] & 78.9 [58.0, 93.9] & 87.7 [83.9, 91.3] \\
\end{longtable}
\normalsize
\footnotesize
\setlength{\tabcolsep}{4pt}
\begin{longtable}{llcccc}
\caption{\textbf{Warning recall: 15 minutes minimum, 2.5 FA/day validation target.} Test FA/day and recall (\%) [95\% CI]. Cutoffs are fixed across contexts and warning requirements.}\label{tab:warning-b2p5-l15}\\
\toprule
Cohort & Forecaster & Test FA/day & Overall & Post-bolus & Non-post-bolus\\\midrule
\endfirsthead
\multicolumn{6}{c}{\tablename\ \thetable\ (continued)}\\\toprule
Cohort & Forecaster & Test FA/day & Overall & Post-bolus & Non-post-bolus\\\midrule
\endhead
\bottomrule\endfoot
DCLP3 ($n=24$) & ARIMAX & 3.21 [2.32, 4.61] & 88.9 [85.9, 91.9] & 72.9 [48.7, 92.9] & 89.1 [86.2, 92.1] \\
 & DLinear & 3.18 [2.37, 4.35] & 91.5 [89.1, 94.0] & 86.6 [66.6, 100.0] & 91.4 [89.0, 94.0] \\
 & GRU & 3.26 [2.43, 4.44] & 92.8 [90.6, 95.0] & 81.7 [61.8, 96.4] & 92.9 [90.8, 95.0] \\
 & iTransformer & 3.22 [2.29, 4.63] & 82.9 [77.5, 87.8] & 63.8 [36.5, 86.5] & 83.4 [77.9, 88.3] \\
 & PatchMLP & 3.04 [2.34, 4.02] & 89.2 [86.0, 92.4] & 70.3 [46.8, 90.9] & 89.8 [86.6, 92.7] \\
 & AttenGluco & 3.21 [2.29, 4.61] & 85.1 [76.4, 90.7] & 66.9 [42.6, 88.3] & 85.7 [76.9, 91.5] \\
 & Glimmer\textsuperscript{*} & 3.09 [2.39, 4.06] & 90.9 [88.3, 93.4] & 80.4 [61.0, 95.2] & 91.0 [88.4, 93.5] \\
 & IOB-linear & 3.17 [2.39, 4.34] & 93.1 [91.1, 95.2] & 86.6 [66.6, 100.0] & 93.1 [91.0, 95.2] \\
\end{longtable}
\normalsize
\footnotesize
\setlength{\tabcolsep}{4pt}
\begin{longtable}{llcccc}
\caption{\textbf{Warning recall: 15 minutes minimum, 3 FA/day validation target.} Test FA/day and recall (\%) [95\% CI]. Cutoffs are fixed across contexts and warning requirements.}\label{tab:warning-b3-l15}\\
\toprule
Cohort & Forecaster & Test FA/day & Overall & Post-bolus & Non-post-bolus\\\midrule
\endfirsthead
\multicolumn{6}{c}{\tablename\ \thetable\ (continued)}\\\toprule
Cohort & Forecaster & Test FA/day & Overall & Post-bolus & Non-post-bolus\\\midrule
\endhead
\bottomrule\endfoot
DCLP3 ($n=24$) & ARIMAX & 3.62 [2.70, 5.00] & 89.4 [86.6, 92.1] & 83.3 [62.1, 100.0] & 89.5 [86.6, 92.3] \\
 & DLinear & 3.79 [2.78, 5.37] & 91.5 [88.4, 94.3] & 92.6 [84.0, 100.0] & 91.4 [88.1, 94.3] \\
 & GRU & 3.72 [2.89, 4.89] & 88.9 [82.2, 93.9] & 79.0 [59.1, 95.5] & 89.3 [82.7, 94.2] \\
 & iTransformer & 3.65 [2.71, 5.06] & 84.6 [79.1, 89.3] & 65.9 [38.6, 88.6] & 85.1 [79.5, 89.7] \\
 & PatchMLP & 3.58 [2.75, 4.75] & 88.3 [85.5, 91.0] & 68.6 [45.9, 88.3] & 88.8 [86.0, 91.5] \\
 & AttenGluco & 3.64 [2.71, 5.03] & 85.2 [76.2, 91.2] & 74.5 [50.2, 94.4] & 85.2 [76.3, 91.4] \\
 & Glimmer\textsuperscript{*} & 3.58 [2.77, 4.74] & 87.3 [78.3, 93.3] & 81.7 [61.8, 96.4] & 87.4 [78.4, 93.4] \\
 & IOB-linear & 3.81 [2.81, 5.38] & 91.5 [88.4, 94.2] & 87.9 [66.7, 100.0] & 91.2 [88.1, 94.1] \\
\end{longtable}
\normalsize
\footnotesize
\setlength{\tabcolsep}{4pt}
\begin{longtable}{llcccc}
\caption{\textbf{Paired recall differences (post-bolus minus non-post-bolus).} Percentage points [95\% CI] among patients with events in both slices, at the 2/day validation target.}\label{tab:paired-differences}\\
\toprule
Cohort & Forecaster & 10 minutes & 15 minutes & 20 minutes & 30 minutes\\\midrule
\endfirsthead
\multicolumn{6}{c}{\tablename\ \thetable\ (continued)}\\\toprule
Cohort & Forecaster & 10 minutes & 15 minutes & 20 minutes & 30 minutes\\\midrule
\endhead
\bottomrule\endfoot
DCLP3 & ARIMAX & 3.2 [0.8, 5.5] & -22.4 [-46.2, -1.7] & -10.8 [-36.0, 12.7] & -10.1 [-29.7, 12.9] \\
DCLP3 & DLinear & 2.4 [0.2, 4.9] & -3.4 [-24.6, 10.0] & -15.1 [-39.4, 7.1] & -10.8 [-31.0, 11.0] \\
DCLP3 & GRU & 1.9 [-2.4, 5.7] & -19.8 [-43.1, 0.5] & -17.9 [-41.6, 2.9] & -9.4 [-29.1, 11.1] \\
DCLP3 & iTransformer & -15.1 [-40.3, 3.0] & -20.4 [-45.6, 1.4] & -25.0 [-47.1, -3.4] & -3.3 [-23.1, 18.3] \\
DCLP3 & PatchMLP & 1.6 [-0.2, 4.0] & -15.7 [-41.4, 5.2] & -26.9 [-49.9, -5.1] & -9.7 [-27.5, 11.6] \\
DCLP3 & AttenGluco & -0.1 [-6.5, 4.9] & -21.3 [-44.6, -0.7] & -21.8 [-47.2, 4.3] & -12.4 [-31.9, 9.1] \\
DCLP3 & Glimmer\textsuperscript{*} & 2.6 [-0.2, 5.1] & -18.4 [-43.8, 2.3] & -19.4 [-43.9, 2.8] & -8.9 [-31.6, 13.6] \\
DCLP3 & IOB-linear & 2.0 [0.6, 3.3] & -8.3 [-29.3, 7.7] & -12.6 [-37.1, 8.8] & -9.6 [-29.9, 12.9] \\
\end{longtable}
\normalsize
\footnotesize
\setlength{\tabcolsep}{4pt}
\begin{longtable}{llcccc}
\caption{\textbf{Other event contexts at 15 minutes minimum warning.} Recall (\%) [95\% CI], 2/day validation target. Meal annotations are available only for PEDAP.}\label{tab:other-warning-contexts}\\
\toprule
Cohort & Forecaster & Nocturnal & Daytime & Post-meal & Other meal contexts\\\midrule
\endfirsthead
\multicolumn{6}{c}{\tablename\ \thetable\ (continued)}\\\toprule
Cohort & Forecaster & Nocturnal & Daytime & Post-meal & Other meal contexts\\\midrule
\endhead
\bottomrule\endfoot
DCLP3 & ARIMAX & 81.5 [70.0, 91.6] & 85.7 [81.6, 89.8] & --- & --- \\
DCLP3 & DLinear & 80.9 [69.7, 90.8] & 94.0 [91.4, 96.3] & --- & --- \\
DCLP3 & GRU & 82.7 [72.0, 91.3] & 90.9 [88.4, 93.5] & --- & --- \\
DCLP3 & iTransformer & 82.1 [70.0, 92.5] & 79.2 [73.4, 84.5] & --- & --- \\
DCLP3 & PatchMLP & 81.9 [70.0, 91.9] & 88.5 [83.7, 92.4] & --- & --- \\
DCLP3 & AttenGluco & 80.7 [67.9, 91.5] & 90.3 [87.2, 93.2] & --- & --- \\
DCLP3 & Glimmer\textsuperscript{*} & 80.4 [66.8, 91.8] & 91.4 [89.0, 93.9] & --- & --- \\
DCLP3 & IOB-linear & 82.0 [70.8, 92.0] & 92.9 [90.3, 95.4] & --- & --- \\
\end{longtable}
\normalsize
\footnotesize
\setlength{\tabcolsep}{4pt}
\begin{longtable}{llrrrrr}
\caption{\textbf{Validation-selected alarm cutoffs.} Cutoffs in mg/dL; the observed low-event threshold remains 70 mg/dL.}\label{tab:alarm-cutoffs}\\
\toprule
Cohort & Forecaster & Target 1 & Target 1.5 & Target 2 & Target 2.5 & Target 3\\\midrule
\endfirsthead
\multicolumn{7}{c}{\tablename\ \thetable\ (continued)}\\\toprule
Cohort & Forecaster & Target 1 & Target 1.5 & Target 2 & Target 2.5 & Target 3\\\midrule
\endhead
\bottomrule\endfoot
DCLP3 & ARIMAX & 81.90 & 87.35 & 91.65 & 95.60 & 98.45 \\
DCLP3 & DLinear & 78.05 & 83.10 & 87.85 & 91.70 & 94.75 \\
DCLP3 & GRU & 79.00 & 84.30 & 89.00 & 92.65 & 95.70 \\
DCLP3 & iTransformer & 78.55 & 83.35 & 87.15 & 90.65 & 93.70 \\
DCLP3 & PatchMLP & 74.80 & 78.85 & 83.25 & 86.75 & 89.55 \\
DCLP3 & AttenGluco & 77.30 & 82.30 & 86.75 & 90.05 & 92.95 \\
DCLP3 & Glimmer\textsuperscript{*} & 74.55 & 79.30 & 83.80 & 87.45 & 90.10 \\
DCLP3 & IOB-linear & 78.50 & 83.80 & 88.40 & 92.30 & 95.40 \\
\end{longtable}
\normalsize

\clearpage\begin{figure}[p]\centering\includegraphics[width=\textwidth,height=.86\textheight,keepaspectratio]{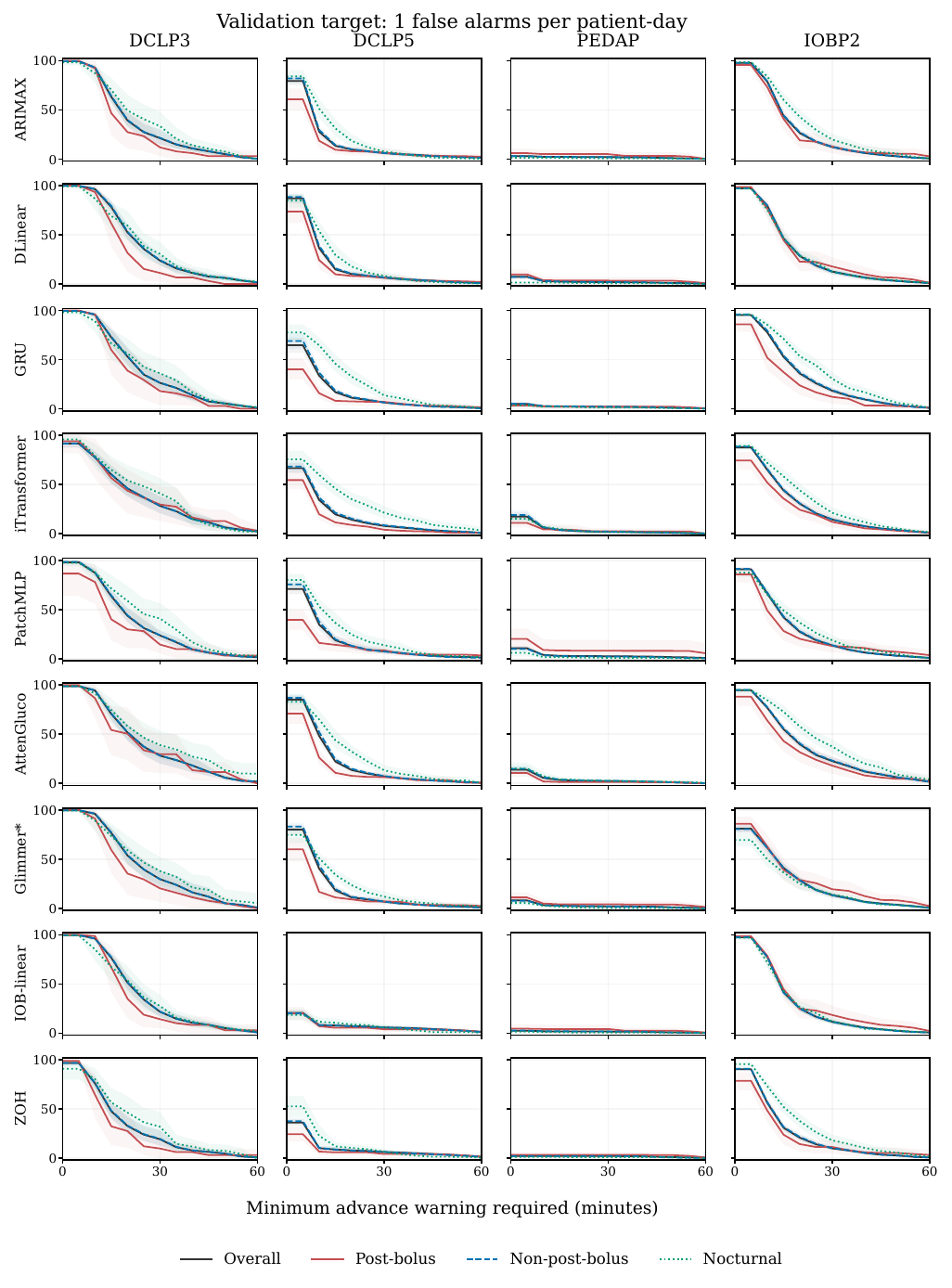}\caption{\textbf{All-model warning curves at a 1 FA/day validation target.} Cutoffs are frozen for test and shared across event contexts. Bands are pointwise 95\% patient-bootstrap intervals. Actual rates and event support are tabulated separately.}\end{figure}
\clearpage\begin{figure}[p]\centering\includegraphics[width=\textwidth,height=.86\textheight,keepaspectratio]{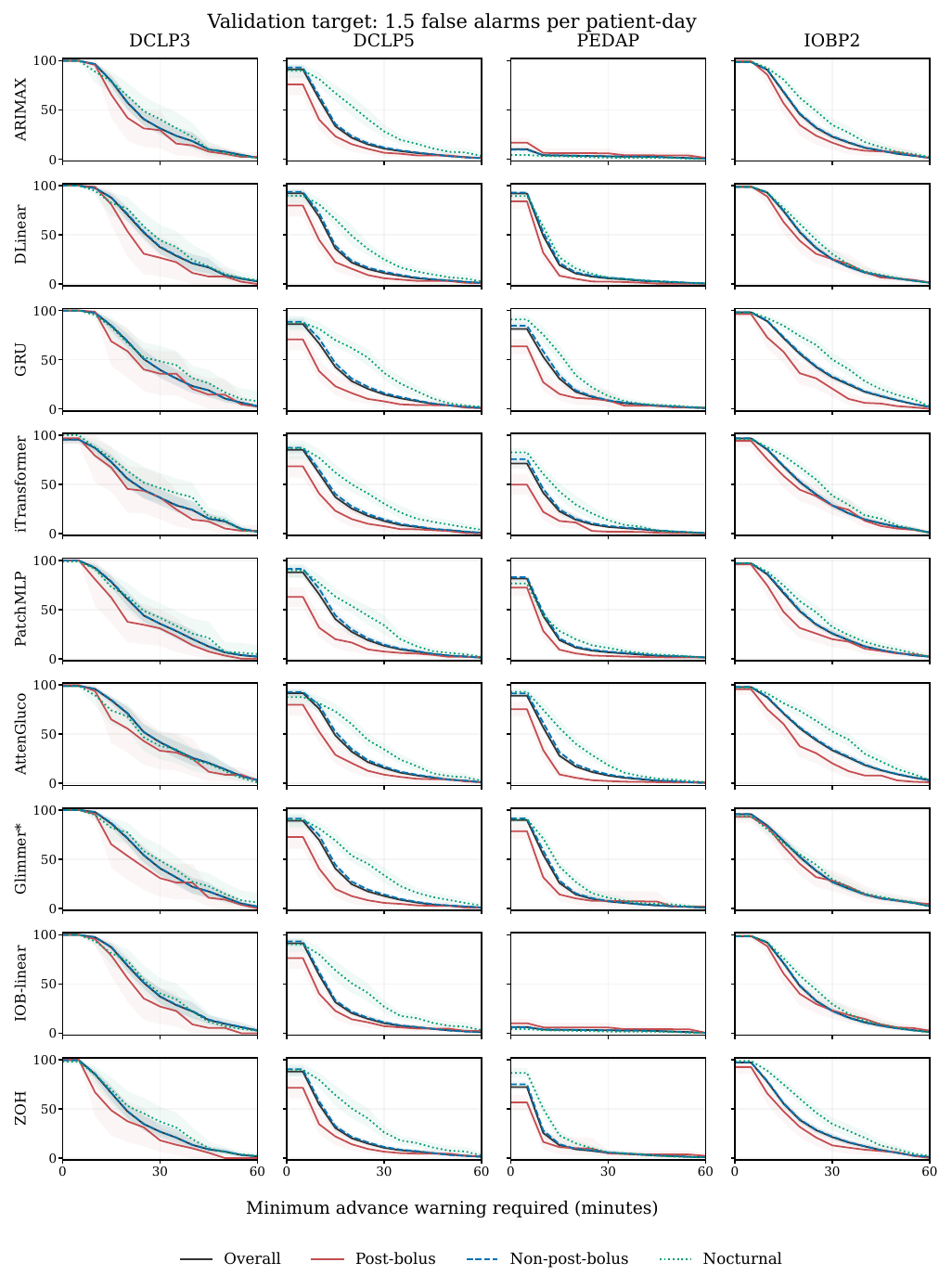}\caption{\textbf{All-model warning curves at a 1.5 FA/day validation target.} Cutoffs are frozen for test and shared across event contexts. Bands are pointwise 95\% patient-bootstrap intervals. Actual rates and event support are tabulated separately.}\end{figure}
\clearpage\begin{figure}[p]\centering\includegraphics[width=\textwidth,height=.86\textheight,keepaspectratio]{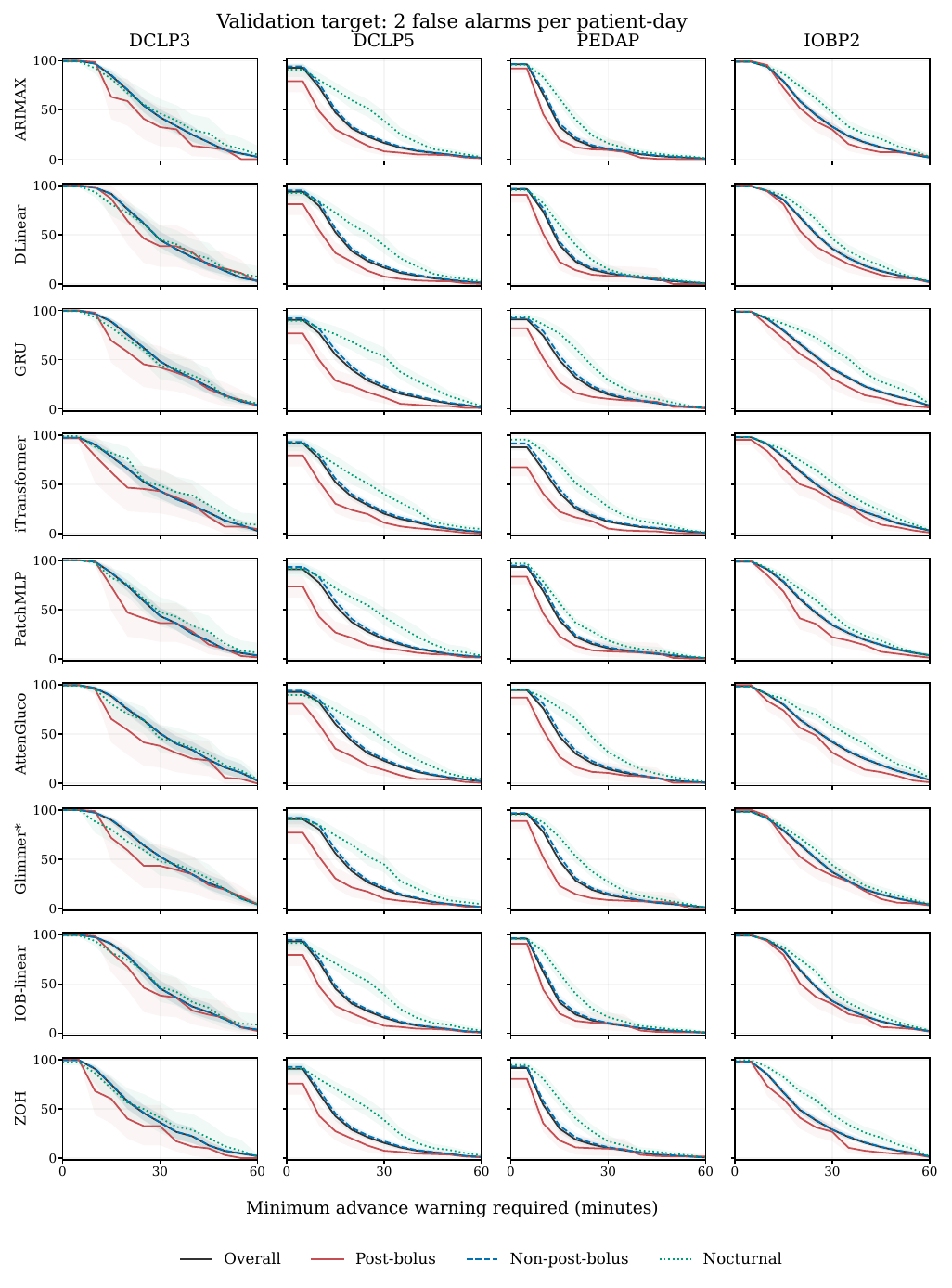}\caption{\textbf{All-model warning curves at a 2 FA/day validation target.} Cutoffs are frozen for test and shared across event contexts. Bands are pointwise 95\% patient-bootstrap intervals. Actual rates and event support are tabulated separately.}\end{figure}
\clearpage\begin{figure}[p]\centering\includegraphics[width=\textwidth,height=.86\textheight,keepaspectratio]{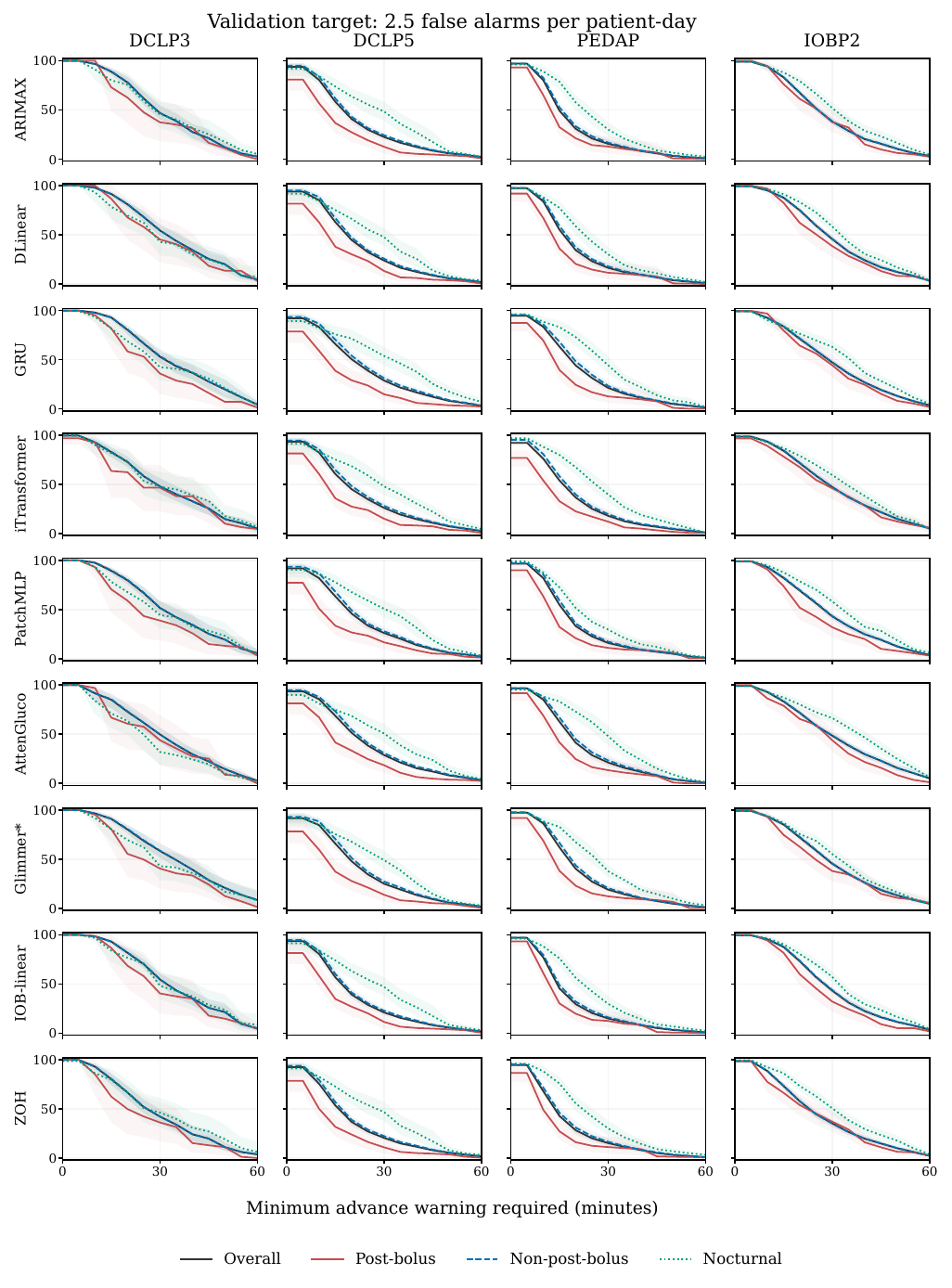}\caption{\textbf{All-model warning curves at a 2.5 FA/day validation target.} Cutoffs are frozen for test and shared across event contexts. Bands are pointwise 95\% patient-bootstrap intervals. Actual rates and event support are tabulated separately.}\end{figure}
\clearpage\begin{figure}[p]\centering\includegraphics[width=\textwidth,height=.86\textheight,keepaspectratio]{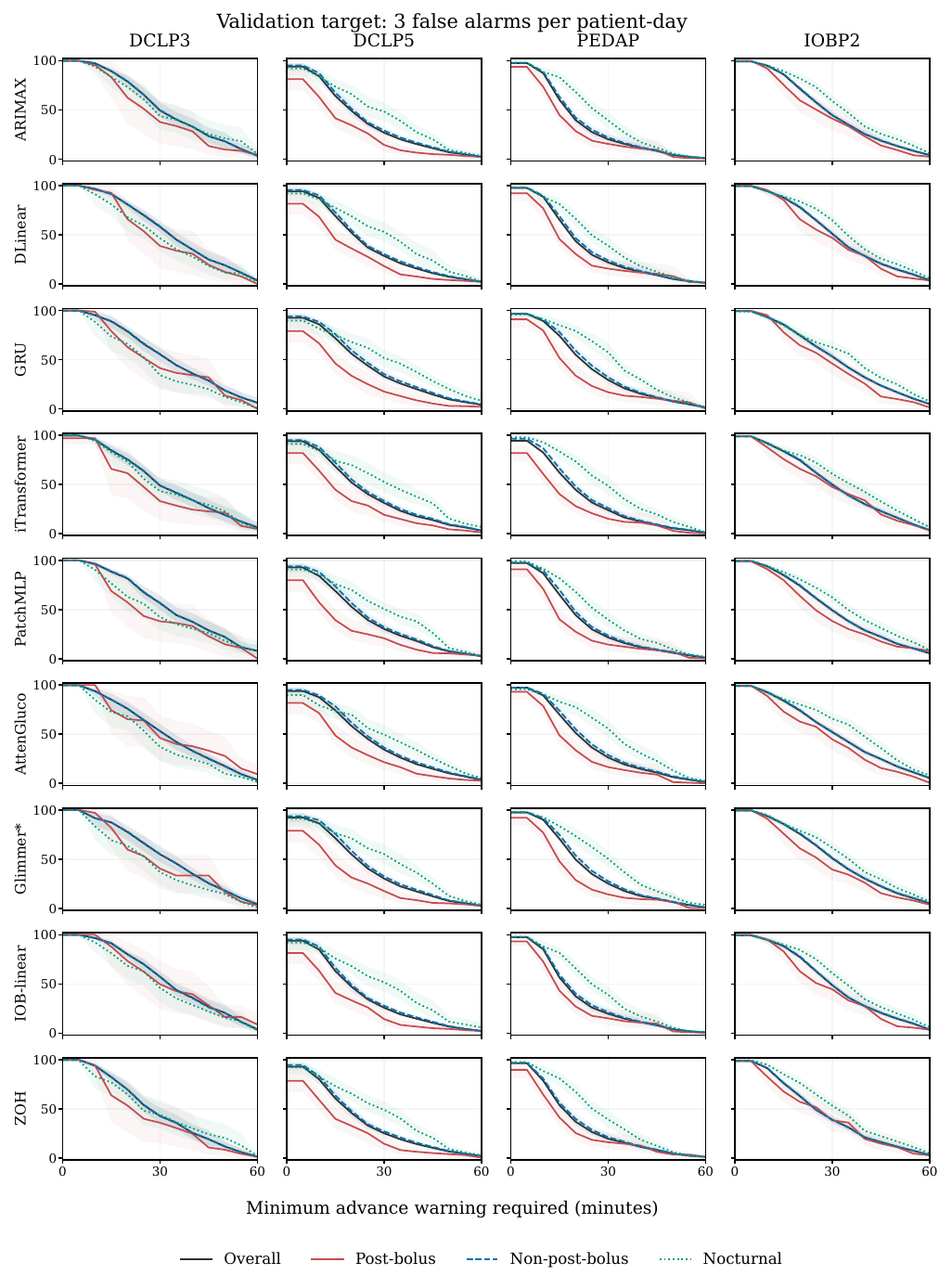}\caption{\textbf{All-model warning curves at a 3 FA/day validation target.} Cutoffs are frozen for test and shared across event contexts. Bands are pointwise 95\% patient-bootstrap intervals. Actual rates and event support are tabulated separately.}\end{figure}

\FloatBarrier

\section{Alarm-Definition Sensitivity}
\label{app:sensitivity}
These sensitivity analyses vary one definition at a time using 30-minute
forecasts, a fixed 70~mg/dL alarm cutoff, and full-timeline matching. They are
a separate protocol from the 60-minute, validation-calibrated warning curves;
their event sets, rates, and recall must not be combined with those curves.
The post-bolus window varies over 15, 30, 45, and 60 minutes; minimum low-event
sample counts correspond to nominal 10, 15, 20, and 30-minute definitions; and
refractory periods vary over 15, 30, and 60 minutes. Slices with fewer events can have
wide intervals, and recall does not change monotonically across all cohorts.
\footnotesize
\setlength{\tabcolsep}{4pt}
\begin{longtable}{lrrrrcc}
\caption{\textbf{Definition sensitivity for Glimmer\textsuperscript{*}.} Full-timeline matching with 30-minute forecasts and a fixed 70 mg/dL alarm cutoff; this recall has no 15-minute minimum-warning requirement.}\label{tab:alarm-sensitivity}\\
\toprule
Cohort & Bolus window & Low duration & Refractory & Events & Recall (\%) [95\% CI] & FA/day [95\% CI]\\\midrule
\endfirsthead
\multicolumn{7}{c}{\tablename\ \thetable\ (continued)}\\\toprule
Cohort & Bolus window & Low duration & Refractory & Events & Recall (\%) [95\% CI] & FA/day [95\% CI]\\\midrule
\endhead
\bottomrule\endfoot
DCLP3 & 15 & 15 & 30 & 13 & 100.0 [100.0, 100.0] & 2.02 [1.46, 2.79] \\
DCLP3 & 30 & 10 & 30 & 45 & 98.1 [93.8, 100.0] & 1.96 [1.40, 2.73] \\
DCLP3 & 30 & 15 & 15 & 37 & 97.2 [90.5, 100.0] & 3.41 [2.50, 4.56] \\
DCLP3 & 30 & 15 & 30 & 37 & 97.2 [90.5, 100.0] & 2.02 [1.46, 2.79] \\
DCLP3 & 30 & 15 & 60 & 37 & 88.9 [75.0, 100.0] & 1.28 [0.97, 1.65] \\
DCLP3 & 30 & 20 & 30 & 30 & 95.8 [85.7, 100.0] & 2.08 [1.52, 2.85] \\
DCLP3 & 30 & 30 & 30 & 19 & 100.0 [100.0, 100.0] & 2.20 [1.63, 2.97] \\
DCLP3 & 45 & 15 & 30 & 65 & 98.0 [93.3, 100.0] & 2.02 [1.46, 2.79] \\
DCLP3 & 60 & 15 & 30 & 100 & 98.3 [96.1, 100.0] & 2.02 [1.46, 2.79] \\
\midrule
DCLP5 & 15 & 15 & 30 & 148 & 76.6 [60.8, 91.3] & 2.94 [2.51, 3.39] \\
DCLP5 & 30 & 10 & 30 & 415 & 81.1 [71.5, 90.1] & 2.76 [2.36, 3.18] \\
DCLP5 & 30 & 15 & 15 & 335 & 80.3 [70.3, 90.0] & 5.13 [4.34, 6.00] \\
DCLP5 & 30 & 15 & 30 & 335 & 80.3 [70.3, 90.0] & 2.94 [2.51, 3.39] \\
DCLP5 & 30 & 15 & 60 & 335 & 74.9 [65.7, 84.0] & 1.95 [1.68, 2.25] \\
DCLP5 & 30 & 20 & 30 & 257 & 80.4 [70.5, 90.0] & 3.14 [2.67, 3.63] \\
DCLP5 & 30 & 30 & 30 & 113 & 67.1 [48.7, 84.5] & 3.50 [2.97, 4.06] \\
DCLP5 & 45 & 15 & 30 & 560 & 87.0 [79.8, 93.4] & 2.94 [2.51, 3.39] \\
DCLP5 & 60 & 15 & 30 & 752 & 88.3 [81.2, 94.4] & 2.94 [2.51, 3.39] \\
\midrule
PEDAP & 15 & 15 & 30 & 204 & 81.1 [62.6, 96.0] & 2.57 [2.26, 2.91] \\
PEDAP & 30 & 10 & 30 & 607 & 91.0 [85.1, 95.4] & 2.34 [2.05, 2.64] \\
PEDAP & 30 & 15 & 15 & 486 & 91.2 [85.3, 95.7] & 4.50 [3.93, 5.11] \\
PEDAP & 30 & 15 & 30 & 486 & 91.1 [85.3, 95.6] & 2.57 [2.26, 2.91] \\
PEDAP & 30 & 15 & 60 & 486 & 82.6 [71.3, 90.7] & 1.78 [1.56, 2.00] \\
PEDAP & 30 & 20 & 30 & 352 & 88.6 [80.3, 94.8] & 2.81 [2.47, 3.16] \\
PEDAP & 30 & 30 & 30 & 154 & 88.6 [79.8, 95.3] & 3.19 [2.82, 3.58] \\
PEDAP & 45 & 15 & 30 & 764 & 92.7 [88.8, 95.9] & 2.57 [2.26, 2.91] \\
PEDAP & 60 & 15 & 30 & 1062 & 93.3 [89.1, 96.3] & 2.57 [2.26, 2.91] \\
\midrule
IOBP2 & 15 & 15 & 30 & 57 & 95.7 [90.4, 100.0] & 2.04 [1.83, 2.26] \\
IOBP2 & 30 & 10 & 30 & 143 & 95.6 [90.8, 99.0] & 1.93 [1.73, 2.14] \\
IOBP2 & 30 & 15 & 15 & 121 & 97.4 [94.7, 99.6] & 3.58 [3.19, 4.00] \\
IOBP2 & 30 & 15 & 30 & 121 & 95.0 [89.7, 98.8] & 2.04 [1.83, 2.26] \\
IOBP2 & 30 & 15 & 60 & 121 & 80.8 [71.3, 88.4] & 1.33 [1.21, 1.45] \\
IOBP2 & 30 & 20 & 30 & 96 & 94.4 [88.3, 98.7] & 2.16 [1.94, 2.39] \\
IOBP2 & 30 & 30 & 30 & 46 & 94.6 [85.9, 100.0] & 2.39 [2.15, 2.64] \\
IOBP2 & 45 & 15 & 30 & 276 & 98.2 [96.5, 99.4] & 2.04 [1.83, 2.26] \\
IOBP2 & 60 & 15 & 30 & 465 & 98.3 [96.8, 99.5] & 2.04 [1.83, 2.26] \\
\midrule
\end{longtable}
\normalsize

\FloatBarrier

\section{Simulator Construction and Metric Definitions}
\label{app:simulator}
We use 100 adult virtual subjects split 70/10/20. Ordinary Control-IQ
trajectories span 15 days per subject. Four perturbation onsets on day 15
produce 80 test episodes from 20 held-out subjects; each episode spans four
hours after onset. Subject, random seed, history, and future meals are shared
within an episode. The factual controller-generated future plan is replayed;
counterfactual insulin plans are predetermined modifications, not a new
closed-loop controller adapting to each changed trajectory.

Basal perturbations multiply the future basal sequence for the entire episode
by $\{0,0.5,1.5,2,2.5,3,3.5,4\}$. Bolus perturbations add
$\{0,0.25,0.5,\ldots,3\}$ times the subject's mean bolus at onset. Effect
RMSE and sign agreement compare these eight/thirteen candidates with the
factual episode. Ranking and action selection also include a factual candidate
label. Zero added bolus
and the factual plan are equivalent deliveries but remain distinct labels.

\paragraph{Effect Metrics.}
At each endpoint, eRMSE is the square root of mean squared error in the paired
effects in Eq.~\ref{eq:effects}. Sign agreement is
\begin{equation}
\mathrm{SA}(k)=\frac{1}{N}\sum_{n=1}^{N}
\mathbf 1\{\operatorname{sign}(\Delta^{\mathrm{pred}}_{n,t+k})
=\operatorname{sign}(\Delta^{\mathrm{true}}_{n,t+k})\}.
\end{equation}
Kendall's $\tau_b$ compares simulated and predicted endpoint action orderings
within each episode, accounting for ties \citep{kendall1945ties}:
\begin{equation}
\tau_b=\frac{C-D}{\sqrt{(C+D+T_g)(C+D+T_{\hat g})}},
\end{equation}
where $C$ and $D$ count concordant and discordant action pairs, and $T_g$ and
$T_{\hat g}$ count pairs tied only in simulated and predicted glucose,
respectively. Direction and
ranking are dashed when the model has no action-dependent endpoint response;
eRMSE is still defined. Within each subject, eRMSE and SA pool paired endpoints
across episodes and actions, while Kendall's $\tau_b$ averages episode-level
rankings. We then average subject metrics, and 95\% intervals resample subjects.
Rounded ties in displayed means do not
establish equal performance or statistical superiority.

\paragraph{BGRI Cost and Action Labels.}
The action cost is
\begin{equation}
\begin{aligned}
f(g)&=1.509[(\ln g)^{1.084}-5.381],\\
J(a)&=\frac{1}{48}\sum_{k=1}^{48}10 f\bigl(g_{t+k}(a)\bigr)^2.
\end{aligned}
\label{eq:bgri}
\end{equation}
Simulator and predicted glucose are clipped below at 1~mg/dL before the
transform, making the cost finite even for nonphysiological predictions.
There is no additional terminal penalty or evaluation beyond four hours.
The oracle chooses the lowest true cost, breaking exact ties toward the lower
action scale with source-order resolution for duplicate labels. Model choice
uses the lowest predicted cost in deterministic scale order; the factual
label receives a zero sorting key, and equal keys retain column order.
This preserves the same rule across models but makes exact-label AMR sensitive
to ties and equivalent actions. Regret compares actual costs and is therefore
reported alongside AMR. The retrospective constant minimizes test average
regret within each menu; its interval conditions on the selected action and
does not incorporate the optimism of test-set selection.
\FloatBarrier

\section{Additional Simulator Results}
\label{app:sim-results}
The extended endpoint tables use the same selected simulator models for both
intervention families. The restricted-menu analysis recomputes both the oracle
and model choice using scale parameters in $[0,2]$, retaining the factual plan.
The basal menu then has five labels and the bolus menu ten. A $2\times$ added
bolus is not a bound on total insulin delivery and does not establish safety.
\footnotesize
\setlength{\tabcolsep}{4pt}
\begin{longtable}{llccc}
\caption{\textbf{Insulin-response prediction at 30 minutes.} Subject means [95\% CI]; SA is percent. DLinear has no action-dependent endpoint response.}\label{tab:sim-endpoint-30}\\
\toprule
Intervention & Forecaster & eRMSE (mg/dL) & SA (\%) & Kendall $\tau_b$\\\midrule
\endfirsthead
\multicolumn{5}{c}{\tablename\ \thetable\ (continued)}\\\toprule
Intervention & Forecaster & eRMSE (mg/dL) & SA (\%) & Kendall $\tau_b$\\\midrule
\endhead
\bottomrule\endfoot
Basal & ARIMAX & 52.77 [43.38, 65.75] & 5.5 [1.9, 9.7] & -0.84 [-0.92, -0.75] \\
Basal & DLinear & 0.06 [0.04, 0.09] & --- & --- \\
Basal & GRU & 15.18 [13.53, 16.90] & 13.9 [8.9, 19.5] & -0.66 [-0.77, -0.55] \\
Basal & iTransformer & 26.10 [22.97, 29.39] & 8.0 [4.2, 11.7] & -0.75 [-0.84, -0.66] \\
Basal & PatchMLP & 22.71 [20.60, 25.22] & 6.1 [2.3, 10.2] & -0.82 [-0.90, -0.73] \\
Basal & AttenGluco & 19.65 [17.86, 21.45] & 12.5 [7.7, 17.2] & -0.67 [-0.77, -0.57] \\
Basal & Glimmer\textsuperscript{*} & 21.83 [19.87, 24.09] & 6.6 [3.1, 10.5] & -0.79 [-0.88, -0.70] \\
\midrule
Bolus & ARIMAX & 64.13 [55.67, 73.24] & 14.5 [11.7, 17.5] & -0.99 [-1.00, -0.99] \\
Bolus & DLinear & 1.00 [0.70, 1.40] & --- & --- \\
Bolus & GRU & 30.96 [28.88, 33.21] & 12.2 [10.0, 14.8] & -0.88 [-0.90, -0.85] \\
Bolus & iTransformer & 41.54 [37.93, 45.06] & 7.4 [6.4, 8.6] & -0.96 [-0.96, -0.95] \\
Bolus & PatchMLP & 32.26 [29.47, 34.93] & 7.6 [6.6, 8.7] & -0.94 [-0.95, -0.93] \\
Bolus & AttenGluco & 6.76 [5.87, 7.83] & 26.3 [20.4, 32.0] & -0.53 [-0.63, -0.42] \\
Bolus & Glimmer\textsuperscript{*} & 33.64 [30.61, 36.61] & 7.7 [6.7, 8.7] & -0.93 [-0.95, -0.91] \\
\midrule
\end{longtable}
\normalsize
\footnotesize
\setlength{\tabcolsep}{4pt}
\begin{longtable}{llccc}
\caption{\textbf{Insulin-response prediction at 60 minutes.} Subject means [95\% CI]; SA is percent. DLinear has no action-dependent endpoint response.}\label{tab:sim-endpoint-60}\\
\toprule
Intervention & Forecaster & eRMSE (mg/dL) & SA (\%) & Kendall $\tau_b$\\\midrule
\endfirsthead
\multicolumn{5}{c}{\tablename\ \thetable\ (continued)}\\\toprule
Intervention & Forecaster & eRMSE (mg/dL) & SA (\%) & Kendall $\tau_b$\\\midrule
\endhead
\bottomrule\endfoot
Basal & ARIMAX & 80.29 [65.98, 97.21] & 1.2 [0.0, 3.6] & -0.97 [-1.00, -0.92] \\
Basal & DLinear & 1.44 [0.98, 1.98] & --- & --- \\
Basal & GRU & 31.65 [28.73, 34.38] & 8.3 [3.9, 13.8] & -0.85 [-0.92, -0.75] \\
Basal & iTransformer & 39.47 [34.98, 44.27] & 5.3 [1.4, 9.8] & -0.85 [-0.92, -0.78] \\
Basal & PatchMLP & 35.89 [32.99, 39.38] & 1.1 [0.0, 2.8] & -0.94 [-0.98, -0.89] \\
Basal & AttenGluco & 34.56 [32.07, 36.87] & 11.1 [6.1, 16.1] & -0.80 [-0.88, -0.71] \\
Basal & Glimmer\textsuperscript{*} & 35.38 [32.60, 38.67] & 1.6 [0.0, 4.1] & -0.92 [-0.97, -0.85] \\
\midrule
Bolus & ARIMAX & 91.88 [81.20, 103.45] & 11.6 [9.4, 14.1] & -0.98 [-0.99, -0.96] \\
Bolus & DLinear & 8.61 [6.76, 10.90] & --- & --- \\
Bolus & GRU & 48.32 [44.11, 52.27] & 17.2 [14.0, 20.5] & -0.82 [-0.86, -0.77] \\
Bolus & iTransformer & 46.53 [42.35, 50.47] & 8.1 [6.9, 9.4] & -0.92 [-0.94, -0.89] \\
Bolus & PatchMLP & 36.42 [33.24, 39.47] & 10.0 [8.0, 12.3] & -0.87 [-0.91, -0.82] \\
Bolus & AttenGluco & 16.46 [14.37, 19.00] & 25.3 [19.9, 30.3] & -0.60 [-0.69, -0.52] \\
Bolus & Glimmer\textsuperscript{*} & 36.24 [33.42, 39.29] & 10.3 [8.2, 12.7] & -0.86 [-0.89, -0.82] \\
\midrule
\end{longtable}
\normalsize
\footnotesize
\setlength{\tabcolsep}{4pt}
\begin{longtable}{llccc}
\caption{\textbf{Insulin-response prediction at 120 minutes.} Subject means [95\% CI]; SA is percent. DLinear has no action-dependent endpoint response.}\label{tab:sim-endpoint-120}\\
\toprule
Intervention & Forecaster & eRMSE (mg/dL) & SA (\%) & Kendall $\tau_b$\\\midrule
\endfirsthead
\multicolumn{5}{c}{\tablename\ \thetable\ (continued)}\\\toprule
Intervention & Forecaster & eRMSE (mg/dL) & SA (\%) & Kendall $\tau_b$\\\midrule
\endhead
\bottomrule\endfoot
Basal & ARIMAX & 106.68 [90.06, 123.88] & 0.0 [0.0, 0.0] & -1.00 [-1.00, -0.99] \\
Basal & DLinear & 13.84 [10.80, 16.85] & --- & --- \\
Basal & GRU & 55.32 [50.30, 59.88] & 3.1 [0.5, 6.1] & -0.95 [-0.99, -0.89] \\
Basal & iTransformer & 61.39 [55.51, 67.59] & 2.3 [0.2, 5.0] & -0.89 [-0.95, -0.83] \\
Basal & PatchMLP & 58.65 [54.21, 63.58] & 0.5 [0.0, 1.2] & -0.96 [-0.98, -0.94] \\
Basal & AttenGluco & 56.37 [52.94, 60.13] & 4.8 [1.4, 8.4] & -0.87 [-0.94, -0.80] \\
Basal & Glimmer\textsuperscript{*} & 58.85 [54.60, 63.21] & 0.6 [0.0, 1.4] & -0.95 [-0.98, -0.92] \\
\midrule
Bolus & ARIMAX & 39.88 [35.81, 44.04] & 13.3 [10.7, 15.9] & -0.81 [-0.86, -0.75] \\
Bolus & DLinear & 27.43 [23.52, 31.75] & --- & --- \\
Bolus & GRU & 41.23 [36.73, 45.85] & 33.7 [28.8, 38.6] & -0.42 [-0.52, -0.33] \\
Bolus & iTransformer & 42.54 [38.68, 46.84] & 18.3 [14.1, 22.5] & -0.54 [-0.64, -0.44] \\
Bolus & PatchMLP & 30.49 [26.66, 34.43] & 45.6 [39.9, 50.9] & -0.08 [-0.20, 0.02] \\
Bolus & AttenGluco & 35.08 [31.25, 39.17] & 33.1 [26.1, 40.0] & -0.41 [-0.52, -0.30] \\
Bolus & Glimmer\textsuperscript{*} & 31.14 [27.40, 34.99] & 42.5 [36.3, 48.2] & -0.09 [-0.22, 0.03] \\
\midrule
\end{longtable}
\normalsize
\footnotesize
\setlength{\tabcolsep}{4pt}
\begin{longtable}{llccc}
\caption{\textbf{Insulin-response prediction at 240 minutes.} Subject means [95\% CI]; SA is percent. DLinear has no action-dependent endpoint response.}\label{tab:sim-endpoint-240}\\
\toprule
Intervention & Forecaster & eRMSE (mg/dL) & SA (\%) & Kendall $\tau_b$\\\midrule
\endfirsthead
\multicolumn{5}{c}{\tablename\ \thetable\ (continued)}\\\toprule
Intervention & Forecaster & eRMSE (mg/dL) & SA (\%) & Kendall $\tau_b$\\\midrule
\endhead
\bottomrule\endfoot
Basal & ARIMAX & 141.32 [129.46, 153.69] & 4.7 [2.0, 8.3] & -0.84 [-0.88, -0.80] \\
Basal & DLinear & 52.10 [46.39, 58.94] & --- & --- \\
Basal & GRU & 86.70 [80.25, 92.88] & 9.5 [4.8, 14.2] & -0.72 [-0.81, -0.65] \\
Basal & iTransformer & 98.09 [90.13, 106.18] & 6.1 [2.3, 10.9] & -0.77 [-0.83, -0.71] \\
Basal & PatchMLP & 100.33 [93.09, 107.64] & 5.9 [2.5, 10.3] & -0.77 [-0.82, -0.72] \\
Basal & AttenGluco & 97.84 [91.46, 104.76] & 5.5 [2.3, 9.5] & -0.83 [-0.87, -0.78] \\
Basal & Glimmer\textsuperscript{*} & 99.82 [93.18, 106.72] & 5.6 [2.5, 9.7] & -0.79 [-0.84, -0.74] \\
\midrule
Bolus & ARIMAX & 25.73 [22.18, 29.03] & 85.7 [82.1, 88.8] & 0.67 [0.61, 0.73] \\
Bolus & DLinear & 37.01 [33.97, 40.56] & --- & --- \\
Bolus & GRU & 29.92 [27.03, 32.58] & 76.2 [72.0, 80.1] & 0.49 [0.41, 0.57] \\
Bolus & iTransformer & 38.07 [34.42, 41.46] & 47.8 [41.3, 53.9] & -0.08 [-0.19, 0.04] \\
Bolus & PatchMLP & 31.07 [27.86, 34.35] & 76.4 [71.7, 81.0] & 0.44 [0.36, 0.53] \\
Bolus & AttenGluco & 38.77 [35.69, 41.64] & 61.7 [55.8, 66.7] & 0.12 [0.01, 0.23] \\
Bolus & Glimmer\textsuperscript{*} & 31.44 [28.55, 34.52] & 73.4 [68.9, 77.9] & 0.43 [0.34, 0.52] \\
\midrule
\end{longtable}
\normalsize
\footnotesize
\setlength{\tabcolsep}{4pt}
\begin{longtable}{llcc}
\caption{\textbf{Full-menu action selection with uncertainty.} Subject means [95\% CI]. AMR is percent; bold: best trained-model mean.}\label{tab:policy-full-ci}\\
\toprule
Selection & Policy & AMR (\%) & Regret\\\midrule
\endfirsthead
\multicolumn{4}{c}{\tablename\ \thetable\ (continued)}\\\toprule
Selection & Policy & AMR (\%) & Regret\\\midrule
\endhead
\bottomrule\endfoot
Basal & \emph{Retrospective constant} & 47.50 [37.50, 57.50] & 0.785 [0.506, 1.083] \\
 & \emph{Control-IQ} & 7.50 [2.50, 12.50] & 3.035 [2.675, 3.403] \\
\cmidrule(l){2-4}
 & ARIMAX & 3.75 [0.00, 7.50] & 4.936 [4.261, 5.714] \\
 & DLinear & 7.50 [2.50, 12.50] & 5.031 [4.324, 5.816] \\
 & GRU & 6.25 [1.25, 12.50] & 4.680 [4.057, 5.337] \\
 & iTransformer & 7.50 [2.50, 12.50] & 4.587 [3.932, 5.281] \\
 & PatchMLP & \textbf{11.25 [5.00, 18.75]} & 4.721 [4.077, 5.391] \\
 & AttenGluco & 7.50 [2.50, 13.75] & \textbf{4.476 [3.871, 5.106]} \\
 & Glimmer\textsuperscript{*} & 5.00 [1.25, 10.00] & 4.745 [4.110, 5.415] \\
\midrule
Bolus & \emph{Retrospective constant} & 47.50 [38.75, 56.28] & 1.150 [0.799, 1.499] \\
 & \emph{Control-IQ} & 3.75 [0.00, 7.50] & 3.904 [3.444, 4.388] \\
\cmidrule(l){2-4}
 & ARIMAX & 6.25 [1.25, 11.25] & 3.848 [3.389, 4.336] \\
 & DLinear & 6.25 [2.50, 11.25] & 3.904 [3.444, 4.388] \\
 & GRU & 8.75 [3.75, 13.75] & 3.062 [2.548, 3.587] \\
 & iTransformer & 11.25 [5.00, 18.75] & 3.659 [3.010, 4.266] \\
 & PatchMLP & 30.00 [22.50, 37.50] & 1.882 [1.434, 2.382] \\
 & AttenGluco & 13.75 [7.50, 20.00] & 1.960 [1.591, 2.382] \\
 & Glimmer\textsuperscript{*} & \textbf{31.25 [23.75, 38.75]} & \textbf{1.715 [1.307, 2.176]} \\
\end{longtable}
\normalsize

\footnotesize
\setlength{\tabcolsep}{4pt}
\begin{longtable}{llcc}
\caption{\textbf{Action selection with a $2\times$ cap.} Model choices and oracle are both restricted; subject means [95\% CI]. The bolus cap limits added, not total, insulin.}\label{tab:restricted-policy}\\
\toprule
Selection & Policy & AMR (\%) & Regret\\\midrule
\endfirsthead
\multicolumn{4}{c}{\tablename\ \thetable\ (continued)}\\\toprule
Selection & Policy & AMR (\%) & Regret\\\midrule
\endhead
\bottomrule\endfoot
Basal & \emph{Retrospective constant} & 75.00 [67.50, 82.50] & 0.219 [0.115, 0.323] \\
 & \emph{Control-IQ} & 7.50 [2.50, 12.50] & 1.547 [1.315, 1.799] \\
\cmidrule(l){2-4}
 & ARIMAX & 3.75 [0.00, 7.50] & 3.442 [2.839, 4.099] \\
 & DLinear & 7.50 [2.50, 12.50] & 3.543 [2.907, 4.216] \\
 & GRU & 6.25 [1.25, 12.50] & 3.181 [2.666, 3.724] \\
 & iTransformer & 7.50 [2.50, 12.50] & 3.093 [2.514, 3.694] \\
 & PatchMLP & \textbf{11.25 [5.00, 18.75]} & 3.212 [2.652, 3.783] \\
 & AttenGluco & 7.50 [2.50, 13.75] & \textbf{2.962 [2.445, 3.501]} \\
 & Glimmer\textsuperscript{*} & 5.00 [1.25, 10.00] & 3.237 [2.715, 3.791] \\
\midrule
Bolus & \emph{Retrospective constant} & 56.25 [47.50, 65.00] & 0.836 [0.578, 1.101] \\
 & \emph{Control-IQ} & 3.75 [0.00, 7.50] & 3.095 [2.715, 3.483] \\
\cmidrule(l){2-4}
 & ARIMAX & 6.25 [1.25, 11.25] & 3.039 [2.659, 3.437] \\
 & DLinear & 6.25 [2.50, 11.25] & 3.095 [2.715, 3.483] \\
 & GRU & 13.75 [7.50, 21.25] & 2.276 [1.835, 2.748] \\
 & iTransformer & 8.75 [2.50, 15.00] & 2.997 [2.464, 3.480] \\
 & PatchMLP & \textbf{30.00 [20.00, 40.00]} & 1.624 [1.237, 2.007] \\
 & AttenGluco & 21.25 [13.75, 27.53] & \textbf{1.431 [1.145, 1.752]} \\
 & Glimmer\textsuperscript{*} & 27.50 [17.50, 37.50] & 1.827 [1.359, 2.371] \\
\end{longtable}
\normalsize

\begin{table}[!htbp]
\centering
\small
\setlength{\tabcolsep}{4pt}
\caption{\textbf{Observed and interventional insulin--glucose associations.}}
\label{tab:confounding}
\begin{tabular}{@{}lrrrr@{}}
\toprule
& \multicolumn{2}{c}{\textbf{Observational}} & \multicolumn{2}{c}{\textbf{Interventional}} \\
\cmidrule(lr){2-3}\cmidrule(lr){4-5}
\textbf{Channel} & $r$ & \textbf{mg/dL/U} & $r$ & \textbf{mg/dL/U} \\
\midrule
Basal & $+0.451$ & $+65.5$ & $-0.931$ & $-11.5$ \\
Bolus & $+0.458$ & $+7.5$  & $-0.904$ & $-3.3$ \\
\bottomrule
\end{tabular}
\end{table}

\par\bigskip\noindent\begin{minipage}{\textwidth}
\centering\footnotesize
\setlength{\tabcolsep}{3pt}
\captionof{table}{\textbf{Basal selections: ARIMAX.} Rows: simulator-optimal labels; columns: model-selected labels. Entries are episode counts. Unselected columns and zero-total rows are omitted. Fact.: factual plan; numbers: action scale.}
\label{tab:confusion-basal-arimax}
\begin{tabular}{lrrrrrr}\toprule
Oracle & 0 & 0.5 & Fact. & 1.5 & 2 & 2.5\\\midrule
0 & 2 & 2 & 0 & 1 & 0 & 1 \\
0.5 & 1 & 1 & 0 & 1 & 0 & 0 \\
Fact. & 3 & 2 & 0 & 1 & 0 & 0 \\
1.5 & 2 & 2 & 0 & 0 & 1 & 0 \\
2 & 4 & 3 & 1 & 0 & 0 & 0 \\
2.5 & 7 & 1 & 0 & 0 & 0 & 0 \\
3 & 3 & 0 & 0 & 0 & 0 & 0 \\
3.5 & 2 & 1 & 0 & 0 & 0 & 0 \\
4 & 36 & 2 & 0 & 0 & 0 & 0 \\
\bottomrule\end{tabular}
\end{minipage}\par\bigskip
\par\bigskip\noindent\begin{minipage}{\textwidth}
\centering\footnotesize
\setlength{\tabcolsep}{3pt}
\captionof{table}{\textbf{Basal selections: DLinear.} Rows: simulator-optimal labels; columns: model-selected labels. Entries are episode counts. Unselected columns and zero-total rows are omitted. Fact.: factual plan; numbers: action scale.}
\label{tab:confusion-basal-dlinear}
\begin{tabular}{lr}\toprule
Oracle & 0\\\midrule
0 & 6 \\
0.5 & 3 \\
Fact. & 6 \\
1.5 & 5 \\
2 & 8 \\
2.5 & 8 \\
3 & 3 \\
3.5 & 3 \\
4 & 38 \\
\bottomrule\end{tabular}
\end{minipage}\par\bigskip
\par\bigskip\noindent\begin{minipage}{\textwidth}
\centering\footnotesize
\setlength{\tabcolsep}{3pt}
\captionof{table}{\textbf{Basal selections: GRU.} Rows: simulator-optimal labels; columns: model-selected labels. Entries are episode counts. Unselected columns and zero-total rows are omitted. Fact.: factual plan; numbers: action scale.}
\label{tab:confusion-basal-gru}
\begin{tabular}{lrrrrr}\toprule
Oracle & 0 & 0.5 & Fact. & 3 & 4\\\midrule
0 & 2 & 1 & 1 & 1 & 1 \\
0.5 & 1 & 1 & 1 & 0 & 0 \\
Fact. & 0 & 4 & 2 & 0 & 0 \\
1.5 & 1 & 3 & 1 & 0 & 0 \\
2 & 0 & 8 & 0 & 0 & 0 \\
2.5 & 3 & 4 & 1 & 0 & 0 \\
3 & 2 & 1 & 0 & 0 & 0 \\
3.5 & 1 & 2 & 0 & 0 & 0 \\
4 & 35 & 3 & 0 & 0 & 0 \\
\bottomrule\end{tabular}
\end{minipage}\par\bigskip
\par\bigskip\noindent\begin{minipage}{\textwidth}
\centering\footnotesize
\setlength{\tabcolsep}{3pt}
\captionof{table}{\textbf{Basal selections: iTransformer.} Rows: simulator-optimal labels; columns: model-selected labels. Entries are episode counts. Unselected columns and zero-total rows are omitted. Fact.: factual plan; numbers: action scale.}
\label{tab:confusion-basal-itransformer}
\begin{tabular}{lrrrrrrr}\toprule
Oracle & 0 & 0.5 & Fact. & 1.5 & 2.5 & 3 & 4\\\midrule
0 & 1 & 0 & 0 & 3 & 1 & 0 & 1 \\
0.5 & 1 & 1 & 0 & 1 & 0 & 0 & 0 \\
Fact. & 1 & 1 & 3 & 1 & 0 & 0 & 0 \\
1.5 & 1 & 1 & 1 & 1 & 0 & 1 & 0 \\
2 & 1 & 5 & 2 & 0 & 0 & 0 & 0 \\
2.5 & 4 & 3 & 1 & 0 & 0 & 0 & 0 \\
3 & 3 & 0 & 0 & 0 & 0 & 0 & 0 \\
3.5 & 1 & 2 & 0 & 0 & 0 & 0 & 0 \\
4 & 26 & 9 & 3 & 0 & 0 & 0 & 0 \\
\bottomrule\end{tabular}
\end{minipage}\par\bigskip
\par\bigskip\noindent\begin{minipage}{\textwidth}
\centering\footnotesize
\setlength{\tabcolsep}{3pt}
\captionof{table}{\textbf{Basal selections: PatchMLP.} Rows: simulator-optimal labels; columns: model-selected labels. Entries are episode counts. Unselected columns and zero-total rows are omitted. Fact.: factual plan; numbers: action scale.}
\label{tab:confusion-basal-patchmlp}
\begin{tabular}{lrrrrrrr}\toprule
Oracle & 0 & 0.5 & Fact. & 1.5 & 2 & 3.5 & 4\\\midrule
0 & 2 & 0 & 1 & 0 & 1 & 1 & 1 \\
0.5 & 2 & 0 & 0 & 1 & 0 & 0 & 0 \\
Fact. & 0 & 1 & 5 & 0 & 0 & 0 & 0 \\
1.5 & 1 & 1 & 1 & 2 & 0 & 0 & 0 \\
2 & 1 & 5 & 2 & 0 & 0 & 0 & 0 \\
2.5 & 7 & 0 & 1 & 0 & 0 & 0 & 0 \\
3 & 2 & 1 & 0 & 0 & 0 & 0 & 0 \\
3.5 & 1 & 2 & 0 & 0 & 0 & 0 & 0 \\
4 & 33 & 5 & 0 & 0 & 0 & 0 & 0 \\
\bottomrule\end{tabular}
\end{minipage}\par\bigskip
\par\bigskip\noindent\begin{minipage}{\textwidth}
\centering\footnotesize
\setlength{\tabcolsep}{3pt}
\captionof{table}{\textbf{Basal selections: AttenGluco.} Rows: simulator-optimal labels; columns: model-selected labels. Entries are episode counts. Unselected columns and zero-total rows are omitted. Fact.: factual plan; numbers: action scale.}
\label{tab:confusion-basal-attengluco}
\begin{tabular}{lrrrrrr}\toprule
Oracle & 0 & 0.5 & Fact. & 2 & 3 & 4\\\midrule
0 & 2 & 0 & 0 & 2 & 2 & 0 \\
0.5 & 1 & 1 & 1 & 0 & 0 & 0 \\
Fact. & 1 & 2 & 3 & 0 & 0 & 0 \\
1.5 & 1 & 1 & 1 & 1 & 0 & 1 \\
2 & 1 & 5 & 2 & 0 & 0 & 0 \\
2.5 & 3 & 4 & 1 & 0 & 0 & 0 \\
3 & 1 & 2 & 0 & 0 & 0 & 0 \\
3.5 & 1 & 2 & 0 & 0 & 0 & 0 \\
4 & 20 & 16 & 2 & 0 & 0 & 0 \\
\bottomrule\end{tabular}
\end{minipage}\par\bigskip
\par\bigskip\noindent\begin{minipage}{\textwidth}
\centering\footnotesize
\setlength{\tabcolsep}{3pt}
\captionof{table}{\textbf{Basal selections: Glimmer\textsuperscript{*}.} Rows: simulator-optimal labels; columns: model-selected labels. Entries are episode counts. Unselected columns and zero-total rows are omitted. Fact.: factual plan; numbers: action scale.}
\label{tab:confusion-basal-patchmlp_glimmer}
\begin{tabular}{lrrrrrrr}\toprule
Oracle & 0 & 0.5 & Fact. & 1.5 & 2 & 3.5 & 4\\\midrule
0 & 2 & 0 & 1 & 0 & 1 & 1 & 1 \\
0.5 & 2 & 0 & 1 & 0 & 0 & 0 & 0 \\
Fact. & 1 & 2 & 2 & 1 & 0 & 0 & 0 \\
1.5 & 1 & 1 & 2 & 0 & 1 & 0 & 0 \\
2 & 1 & 5 & 2 & 0 & 0 & 0 & 0 \\
2.5 & 6 & 1 & 1 & 0 & 0 & 0 & 0 \\
3 & 3 & 0 & 0 & 0 & 0 & 0 & 0 \\
3.5 & 1 & 2 & 0 & 0 & 0 & 0 & 0 \\
4 & 34 & 4 & 0 & 0 & 0 & 0 & 0 \\
\bottomrule\end{tabular}
\end{minipage}\par\bigskip
\par\bigskip\noindent\begin{minipage}{\textwidth}
\centering\footnotesize
\setlength{\tabcolsep}{3pt}
\captionof{table}{\textbf{Bolus selections: ARIMAX.} Rows: simulator-optimal labels; columns: model-selected labels. Entries are episode counts. Unselected columns and zero-total rows are omitted. Fact.: factual plan; numbers: action scale.}
\label{tab:confusion-bolus-arimax}
\begin{tabular}{lrrrrr}\toprule
Oracle & 0 & 0.25 & 0.5 & 0.75 & 1\\\midrule
0 & 4 & 0 & 1 & 0 & 0 \\
Fact. & 2 & 0 & 0 & 1 & 0 \\
0.25 & 5 & 1 & 0 & 1 & 1 \\
0.5 & 2 & 2 & 0 & 0 & 0 \\
1 & 2 & 0 & 0 & 0 & 0 \\
1.25 & 2 & 0 & 0 & 0 & 0 \\
1.5 & 4 & 1 & 0 & 0 & 0 \\
1.75 & 2 & 0 & 0 & 0 & 0 \\
2 & 4 & 0 & 0 & 0 & 0 \\
2.25 & 1 & 0 & 1 & 0 & 0 \\
2.5 & 2 & 1 & 0 & 0 & 0 \\
2.75 & 2 & 0 & 0 & 0 & 0 \\
3 & 35 & 1 & 1 & 1 & 0 \\
\bottomrule\end{tabular}
\end{minipage}\par\bigskip
\par\bigskip\noindent\begin{minipage}{\textwidth}
\centering\footnotesize
\setlength{\tabcolsep}{3pt}
\captionof{table}{\textbf{Bolus selections: DLinear.} Rows: simulator-optimal labels; columns: model-selected labels. Entries are episode counts. Unselected columns and zero-total rows are omitted. Fact.: factual plan; numbers: action scale.}
\label{tab:confusion-bolus-dlinear}
\begin{tabular}{lr}\toprule
Oracle & 0\\\midrule
0 & 5 \\
Fact. & 3 \\
0.25 & 8 \\
0.5 & 4 \\
1 & 2 \\
1.25 & 2 \\
1.5 & 5 \\
1.75 & 2 \\
2 & 4 \\
2.25 & 2 \\
2.5 & 3 \\
2.75 & 2 \\
3 & 38 \\
\bottomrule\end{tabular}
\end{minipage}\par\bigskip
\par\bigskip\noindent\begin{minipage}{\textwidth}
\centering\footnotesize
\setlength{\tabcolsep}{3pt}
\captionof{table}{\textbf{Bolus selections: GRU.} Rows: simulator-optimal labels; columns: model-selected labels. Entries are episode counts. Unselected columns and zero-total rows are omitted. Fact.: factual plan; numbers: action scale.}
\label{tab:confusion-bolus-gru}
\begin{tabular}{lrrrrrrrrrrr}\toprule
Oracle & 0 & 0.25 & 0.5 & 0.75 & 1 & 1.25 & 1.5 & 1.75 & 2.25 & 2.5 & 3\\\midrule
0 & 4 & 0 & 0 & 0 & 0 & 1 & 0 & 0 & 0 & 0 & 0 \\
Fact. & 2 & 0 & 0 & 1 & 0 & 0 & 0 & 0 & 0 & 0 & 0 \\
0.25 & 5 & 0 & 1 & 0 & 1 & 0 & 0 & 1 & 0 & 0 & 0 \\
0.5 & 1 & 2 & 0 & 0 & 0 & 0 & 1 & 0 & 0 & 0 & 0 \\
1 & 1 & 1 & 0 & 0 & 0 & 0 & 0 & 0 & 0 & 0 & 0 \\
1.25 & 2 & 0 & 0 & 0 & 0 & 0 & 0 & 0 & 0 & 0 & 0 \\
1.5 & 4 & 1 & 0 & 0 & 0 & 0 & 0 & 0 & 0 & 0 & 0 \\
1.75 & 1 & 1 & 0 & 0 & 0 & 0 & 0 & 0 & 0 & 0 & 0 \\
2 & 1 & 1 & 0 & 0 & 0 & 0 & 0 & 0 & 0 & 1 & 1 \\
2.25 & 1 & 1 & 0 & 0 & 0 & 0 & 0 & 0 & 0 & 0 & 0 \\
2.5 & 3 & 0 & 0 & 0 & 0 & 0 & 0 & 0 & 0 & 0 & 0 \\
2.75 & 0 & 1 & 0 & 1 & 0 & 0 & 0 & 0 & 0 & 0 & 0 \\
3 & 15 & 3 & 4 & 2 & 5 & 4 & 0 & 0 & 2 & 0 & 3 \\
\bottomrule\end{tabular}
\end{minipage}\par\bigskip
\par\bigskip\noindent\begin{minipage}{\textwidth}
\centering\footnotesize
\setlength{\tabcolsep}{3pt}
\captionof{table}{\textbf{Bolus selections: iTransformer.} Rows: simulator-optimal labels; columns: model-selected labels. Entries are episode counts. Unselected columns and zero-total rows are omitted. Fact.: factual plan; numbers: action scale.}
\label{tab:confusion-bolus-itransformer}
\begin{tabular}{lrrrrrrr}\toprule
Oracle & 0 & 0.25 & 0.5 & 0.75 & 1.75 & 2.25 & 3\\\midrule
0 & 4 & 0 & 0 & 0 & 0 & 1 & 0 \\
Fact. & 1 & 0 & 1 & 0 & 0 & 0 & 1 \\
0.25 & 4 & 0 & 0 & 1 & 2 & 1 & 0 \\
0.5 & 2 & 1 & 0 & 0 & 0 & 0 & 1 \\
1 & 2 & 0 & 0 & 0 & 0 & 0 & 0 \\
1.25 & 1 & 1 & 0 & 0 & 0 & 0 & 0 \\
1.5 & 4 & 0 & 0 & 1 & 0 & 0 & 0 \\
1.75 & 2 & 0 & 0 & 0 & 0 & 0 & 0 \\
2 & 4 & 0 & 0 & 0 & 0 & 0 & 0 \\
2.25 & 1 & 1 & 0 & 0 & 0 & 0 & 0 \\
2.5 & 3 & 0 & 0 & 0 & 0 & 0 & 0 \\
2.75 & 2 & 0 & 0 & 0 & 0 & 0 & 0 \\
3 & 32 & 0 & 0 & 1 & 0 & 0 & 5 \\
\bottomrule\end{tabular}
\end{minipage}\par\bigskip
\par\bigskip\noindent\begin{minipage}{\textwidth}
\centering\footnotesize
\setlength{\tabcolsep}{3pt}
\captionof{table}{\textbf{Bolus selections: PatchMLP.} Rows: simulator-optimal labels; columns: model-selected labels. Entries are episode counts. Unselected columns and zero-total rows are omitted. Fact.: factual plan; numbers: action scale.}
\label{tab:confusion-bolus-patchmlp}
\begin{tabular}{lrrrrrrrrr}\toprule
Oracle & 0 & 0.25 & 0.5 & 0.75 & 1 & 1.25 & 1.5 & 2.25 & 3\\\midrule
0 & 4 & 0 & 0 & 0 & 0 & 0 & 1 & 0 & 0 \\
Fact. & 2 & 0 & 0 & 0 & 0 & 0 & 0 & 0 & 1 \\
0.25 & 5 & 0 & 0 & 0 & 1 & 1 & 0 & 1 & 0 \\
0.5 & 2 & 0 & 1 & 1 & 0 & 0 & 0 & 0 & 0 \\
1 & 2 & 0 & 0 & 0 & 0 & 0 & 0 & 0 & 0 \\
1.25 & 2 & 0 & 0 & 0 & 0 & 0 & 0 & 0 & 0 \\
1.5 & 2 & 1 & 1 & 0 & 0 & 0 & 0 & 0 & 1 \\
1.75 & 2 & 0 & 0 & 0 & 0 & 0 & 0 & 0 & 0 \\
2 & 2 & 0 & 0 & 0 & 0 & 0 & 0 & 0 & 2 \\
2.25 & 1 & 0 & 1 & 0 & 0 & 0 & 0 & 0 & 0 \\
2.5 & 2 & 0 & 0 & 0 & 0 & 0 & 0 & 0 & 1 \\
2.75 & 2 & 0 & 0 & 0 & 0 & 0 & 0 & 0 & 0 \\
3 & 16 & 1 & 0 & 0 & 1 & 0 & 1 & 0 & 19 \\
\bottomrule\end{tabular}
\end{minipage}\par\bigskip
\par\bigskip\noindent\begin{minipage}{\textwidth}
\centering\footnotesize
\setlength{\tabcolsep}{3pt}
\captionof{table}{\textbf{Bolus selections: AttenGluco.} Rows: simulator-optimal labels; columns: model-selected labels. Entries are episode counts. Unselected columns and zero-total rows are omitted. Fact.: factual plan; numbers: action scale.}
\label{tab:confusion-bolus-attengluco}
\begin{tabular}{lrrrrrrrrrrrrr}\toprule
Oracle & 0 & 0.25 & 0.5 & 0.75 & 1 & 1.25 & 1.5 & 1.75 & 2 & 2.25 & 2.5 & 2.75 & 3\\\midrule
0 & 1 & 1 & 1 & 1 & 0 & 1 & 0 & 0 & 0 & 0 & 0 & 0 & 0 \\
Fact. & 2 & 0 & 0 & 0 & 0 & 0 & 0 & 0 & 0 & 0 & 0 & 0 & 1 \\
0.25 & 5 & 0 & 0 & 0 & 1 & 1 & 0 & 0 & 0 & 0 & 0 & 0 & 1 \\
0.5 & 0 & 1 & 1 & 0 & 0 & 0 & 1 & 0 & 0 & 0 & 0 & 0 & 1 \\
1 & 2 & 0 & 0 & 0 & 0 & 0 & 0 & 0 & 0 & 0 & 0 & 0 & 0 \\
1.25 & 2 & 0 & 0 & 0 & 0 & 0 & 0 & 0 & 0 & 0 & 0 & 0 & 0 \\
1.5 & 2 & 1 & 0 & 0 & 0 & 0 & 0 & 0 & 0 & 0 & 1 & 0 & 1 \\
1.75 & 1 & 0 & 0 & 0 & 1 & 0 & 0 & 0 & 0 & 0 & 0 & 0 & 0 \\
2 & 1 & 0 & 0 & 1 & 0 & 1 & 1 & 0 & 0 & 0 & 0 & 0 & 0 \\
2.25 & 1 & 0 & 0 & 0 & 0 & 1 & 0 & 0 & 0 & 0 & 0 & 0 & 0 \\
2.5 & 2 & 0 & 0 & 0 & 0 & 0 & 1 & 0 & 0 & 0 & 0 & 0 & 0 \\
2.75 & 2 & 0 & 0 & 0 & 0 & 0 & 0 & 0 & 0 & 0 & 0 & 0 & 0 \\
3 & 13 & 0 & 0 & 1 & 0 & 0 & 5 & 3 & 3 & 3 & 0 & 1 & 9 \\
\bottomrule\end{tabular}
\end{minipage}\par\bigskip
\par\bigskip\noindent\begin{minipage}{\textwidth}
\centering\footnotesize
\setlength{\tabcolsep}{3pt}
\captionof{table}{\textbf{Bolus selections: Glimmer\textsuperscript{*}.} Rows: simulator-optimal labels; columns: model-selected labels. Entries are episode counts. Unselected columns and zero-total rows are omitted. Fact.: factual plan; numbers: action scale.}
\label{tab:confusion-bolus-patchmlp_glimmer}
\begin{tabular}{lrrrrrrrrrrr}\toprule
Oracle & 0 & 0.25 & 0.5 & 0.75 & 1 & 1.5 & 1.75 & 2 & 2.5 & 2.75 & 3\\\midrule
0 & 4 & 0 & 0 & 0 & 0 & 1 & 0 & 0 & 0 & 0 & 0 \\
Fact. & 2 & 0 & 0 & 0 & 0 & 0 & 0 & 0 & 0 & 0 & 1 \\
0.25 & 4 & 1 & 0 & 1 & 0 & 1 & 0 & 1 & 0 & 0 & 0 \\
0.5 & 1 & 1 & 1 & 0 & 0 & 1 & 0 & 0 & 0 & 0 & 0 \\
1 & 2 & 0 & 0 & 0 & 0 & 0 & 0 & 0 & 0 & 0 & 0 \\
1.25 & 2 & 0 & 0 & 0 & 0 & 0 & 0 & 0 & 0 & 0 & 0 \\
1.5 & 2 & 1 & 1 & 0 & 0 & 0 & 0 & 0 & 0 & 0 & 1 \\
1.75 & 2 & 0 & 0 & 0 & 0 & 0 & 0 & 0 & 0 & 0 & 0 \\
2 & 2 & 0 & 0 & 0 & 0 & 0 & 0 & 0 & 0 & 1 & 1 \\
2.25 & 1 & 0 & 1 & 0 & 0 & 0 & 0 & 0 & 0 & 0 & 0 \\
2.5 & 2 & 0 & 0 & 0 & 0 & 0 & 0 & 0 & 0 & 0 & 1 \\
2.75 & 1 & 0 & 0 & 0 & 1 & 0 & 0 & 0 & 0 & 0 & 0 \\
3 & 14 & 1 & 1 & 0 & 0 & 0 & 1 & 0 & 1 & 1 & 19 \\
\bottomrule\end{tabular}
\end{minipage}\par\bigskip

\FloatBarrier
\end{document}